\documentclass{article} 
\newif\ifarxiv
\arxivtrue
\usepackage{iclr2027_conference,times}
\ifarxiv\iclrfinalcopy\fi
\usepackage[T1]{fontenc}

\usepackage{amsmath,amsfonts,bm}

\def\eqref#1{equation~\ref{#1}}

\def\1{\bm{1}}

\DeclareMathAlphabet{\mathsfit}{\encodingdefault}{\sfdefault}{m}{sl}
\SetMathAlphabet{\mathsfit}{bold}{\encodingdefault}{\sfdefault}{bx}{n}

\usepackage{amsmath,amssymb}
\usepackage{booktabs}
\usepackage{graphicx}
\usepackage{float}
\usepackage{caption}
\usepackage{microtype}
\usepackage{xcolor}
\usepackage{url}
\usepackage{xurl}
\usepackage{hyperref}
\usepackage{etoc}
\usepackage{placeins}
\hypersetup{pdftitle={Channel Selection for Mixture-of-Experts Decoding Needs No Calibration Data}}
\ifarxiv
\hypersetup{colorlinks, linkcolor=red!60!black, citecolor=blue!60!black, urlcolor=blue!60!black,
  pdfauthor={Geunhyoung Kim, Sung Ju Hwang}}
\else
\hypersetup{hidelinks}
\fi
\newcommand{\se}[1]{{\scriptstyle\pm#1}}  

\title{Channel Selection for Mixture-of-Experts\\Decoding Needs No Calibration Data}

\author{Geunhyoung Kim$^{1}$ \quad Sung Ju Hwang$^{1,2}$ \\[1.5pt]
$^{1}$KAIST \quad $^{2}$DeepAuto.ai \\
\texttt{\{skybluejacket, sungju.hwang\}@kaist.ac.kr}}

\begin{document}
\etocdepthtag.toc{main}

\maketitle
\ifarxiv\lhead{}\renewcommand{\headrulewidth}{0pt}\fi  

\begin{abstract}
Mixture-of-Experts (MoE) models are bottlenecked by loading routed-expert weights from memory during interactive decoding at small batch sizes. Because most intermediate channels in a routed expert
contribute little to any single token, channel selectors predict which channels matter and read only
their weights, typically keeping channels above a threshold calibrated on offline data. We show that
channel selection needs no calibration data. First, a calibrated threshold does not fix how much a selector reads,
because the number of channels passing it varies across inputs; Prox with thresholds calibrated for an 18\% read budget
actually reads only 12.9\% on average across our benchmarks.
Second, top-$k$ keeps each token's $k$ highest-scoring channels and needs no calibration data.
At the calibrated budget, a selector's top-$k$ version matches or beats its threshold version on the same score in 40 of 43 comparisons at 13\% and 18\% budgets, including all 23 on generation tasks.
The top-$k$ versions also avoid the repetition-loop collapse that calibrated thresholds cause on Gemma~4.
Calibrating a threshold thus
provides neither of the two things selectors use it for: a known read budget and a consistent accuracy gain at
that budget. We furthermore introduce a reader built entirely from
the model's weights, which screens channels with a ternary sketch and recovers the skipped ones with a
low-rank map. Across three
post-trained MoE models, our reader retains 98.1\% of full-model accuracy at 17.8\% of routed-expert reads,
with Triton kernels running MoE blocks 1.60 to 1.90$\times$ faster than optimized vLLM at batch
one and up to 2.69$\times$ at larger batches on the same three models.
\end{abstract}

\begin{figure}[H]
\centering
\includegraphics[width=\linewidth]{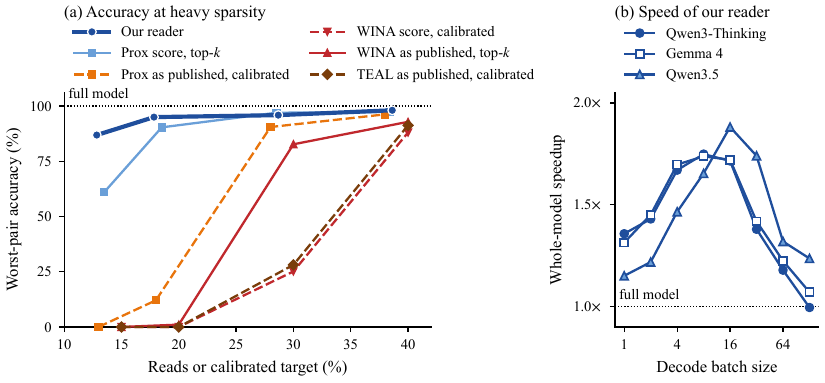}
\caption{\textbf{At heavy sparsity, calibrated selectors lose most of their accuracy on at least one task,
whereas data-free top-$k$ selectors stay accurate and our reader performs best and decodes faster.}
(a)~Worst accuracy over three MoE models and three benchmarks; dashed lines use calibrated thresholds, placed at
the budget they were calibrated for.
See Appendix Figure~\ref{fig:worst-cell-realized} for the same curves with calibrated selectors placed at the reads they actually used. (b)~Whole-model decoding speedup of our reader at 17.8\% of reads over optimized vLLM.}
\label{fig:headline}
\end{figure}

\section{Introduction}
\label{sec:intro}
Mixture-of-Experts (MoE) models scale capacity by routing each token to only a few experts
\citep{shazeer2017outrageously,fedus2022switch}. At the small batch sizes of interactive decoding,
latency is dominated by reading routed-expert weights from GPU memory (DRAM) to on-chip SRAM.
Within each routed expert's gated MLP, however, most intermediate channels contribute little to any
single token and can be skipped with minimal loss \citep{park2026intraexpert}. A channel selector
exploits this by predicting which channels matter and reading only their weights.

Every channel selector couples two components: a channel \emph{score} that estimates each channel's
contribution to the output, and a \emph{keep rule} that decides which channels to compute. To balance
quality and sparsity without retraining, most selectors use a \emph{calibrated threshold}, setting a
fixed per-layer cutoff on offline calibration data so that the model reads a target budget on that
data \citep{lee2024cats,liu2025teal,zhang2025rsparse,liu2026prox}. The alternative keep rule is
\emph{top-$k$}, which keeps each token's $k$ highest-scoring channels and needs no calibration data itself
\citep{liu2025larosa,chen2025wina}. Prior studies disagree on which rule is preferable. Prox
\citep{liu2026prox} argues that a threshold's variable channel count adapts computation to each
token, and both Prox and the Sparsing Law \citep{luo2025sparsing} find their threshold versions ahead of
top-$k$ on dense models. Conversely, other studies find top-$k$ ahead
\citep{liu2025larosa}, while calibration introduces sensitivity to the choice
of calibration data \citep{williams2024calibration,ji2025calibration}.

We show that channel selection for frozen MoE decoding needs no calibration data.
Selectors calibrate a threshold for two reasons, and it delivers neither.
First, a calibrated threshold does not fix how much a selector reads (Section~\ref{sec:drift}).
The number of channels above a fixed cutoff changes with the input, so selectors with calibrated thresholds read less than the budget they were calibrated for in all 72 benchmark runs we made, and calibrating on the models' own reasoning does not fix this.
On Qwen3-Thinking, the count follows the answer written so far more than how much each token needs.
Second, calibration gives no consistent accuracy gain over top-$k$ (Section~\ref{sec:keep-rule}).
We run each selector twice on its own score, once with its calibrated threshold and once with top-$k$, at the budget the threshold was calibrated for.
The top-$k$ version matches or beats the threshold version in 40 of 43 comparisons across four MoE models at 13\% and 18\% budgets.
Under heavy sparsity, Gemma~4 with calibrated thresholds falls into repetition loops during long reasoning, which the same selectors with top-$k$ avoid (Figure~\ref{fig:headline}).

With top-$k$ as the keep rule, performance depends on how well a selector ranks channels before reading them and handles skipped channels.
Our reader is built entirely from the model's weights (Section~\ref{sec:method}).
For each token, it screens all candidate channels with a rotated ternary sketch of the weights (about 7\% of the BF16 checkpoint), computes the top-$k$ exactly, and recovers the skipped channels through a low-rank map of the down projections.

\begin{figure}[!t]
\centering
\includegraphics[width=0.9\linewidth]{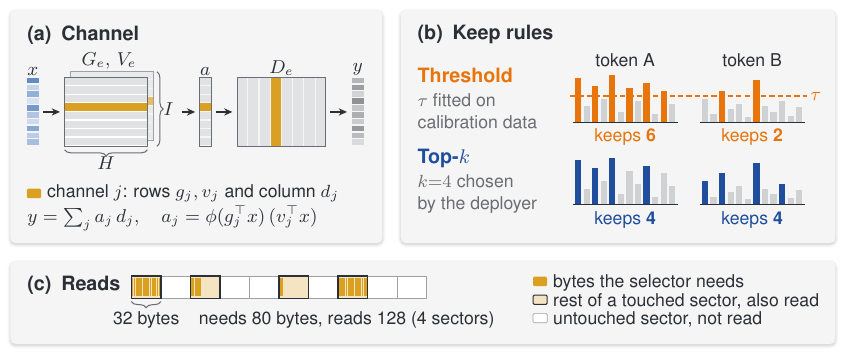}
\caption{\textbf{A channel selector computes only a subset of each routed expert's channels, and hardware
charges every 32-byte sector it touches.}
(a)~Channel $j$ owns row $j$ of the gate and up projections ($G_e,V_e$) and column $j$ of the down projection
($D_e$); $H$ is the hidden width, $I$ the number of channels per expert, $\phi$ the gate activation.
(b)~On the same scores, a calibrated threshold $\tau$ keeps six channels on token~A and two on token~B, while
top-$k$ keeps four on both. (c)~A selector that needs 80 bytes spread over four sectors is charged 128 bytes.}
\label{fig:prelim}
\end{figure}
We evaluate our reader in vLLM \citep{kwon2023vllm} on Qwen3-30B-A3B-Thinking-2507 \citep{yang2025qwen3}, Gemma-4-26B-A4B-it \citep{gemma2026gemma4}, and Qwen3.5-35B-A3B \citep{qwen2026qwen35}, charging every weight and parameter read in 32-byte hardware sectors and checking counts with a GPU profiler (Section~\ref{sec:prelim}).
At both 12.8\% and 17.8\% of routed-expert reads, it is the most accurate tested selector, on average and in its worst model-benchmark pair, which keeps 95.0\% of full-model accuracy at 17.8\%.
On OLMoE \citep{muennighoff2025olmoe}, it matches the full model more closely than nine prior selectors at the budgets at which each was tested (Appendix~\ref{app:olmoe-baselines}).
Our Triton kernels make whole models decode up to 1.88$\times$ faster than optimized vLLM, with the gain peaking at batch size eight or sixteen (Section~\ref{sec:speed}).

\section{Background}
\label{sec:prelim}
In an MoE layer, a router sends each token to a few experts, each a gated MLP \citep{shazeer2020glu}.
We call each intermediate dimension of a routed expert a \emph{channel}: channel $j$ of expert $e$ owns row $j$ of the gate and up projections $G_e$ and $V_e$ and column $j$ of the down projection $D_e$ (Figure~\ref{fig:prelim}a).
The routed output sums all channel outputs, each weighted by its expert's router weight.
In Qwen3-Thinking \citep{yang2025qwen3}, a token goes to 8 experts of 768 channels, so a selector chooses among 6{,}144 channels per layer.
Under a calibrated threshold, the token's scores decide how many channels pass the cutoff; under top-$k$, the deployer fixes this number in advance (Figure~\ref{fig:prelim}b).

We measure cost in \emph{reads}: the bytes loaded per token as a percentage of the full model's routed-expert reads.
Reads include kept-channel weights, sketch or proxy weights, and auxiliary parameters.
Modern NVIDIA GPUs fetch memory in 32-byte sectors \citep{nvidia2025cudabp}, so we charge every touched sector in full (Figure~\ref{fig:prelim}c).
A GPU profiler confirms the count: at batch one on all three post-trained models, it measures cold-cache DRAM reads of 18.3\% against our counted 17.8 to 17.9\% (Appendix~\ref{app:accounting}).

A calibrated threshold is not a selector by itself but a keep rule that selectors such as Prox and TEAL use.
To compare keep rules, we run each such selector twice on the same score: its \emph{threshold version} keeps the calibrated threshold, and its \emph{top-$k$ version} replaces the threshold with top-$k$.
We compare the two at the budget the threshold was calibrated for, because a deployer sets this budget in advance, while the reads the threshold version actually uses, its \emph{realized reads}, are known only afterward (Section~\ref{sec:drift}).
As a check, we also run the top-$k$ version at these realized reads.
Threshold versions read less than their budget, so this check favors them (Section~\ref{sec:keep-rule}).


\section{Related work}
\label{sec:related}
\begin{figure}[t]
\centering
\includegraphics[width=\linewidth]{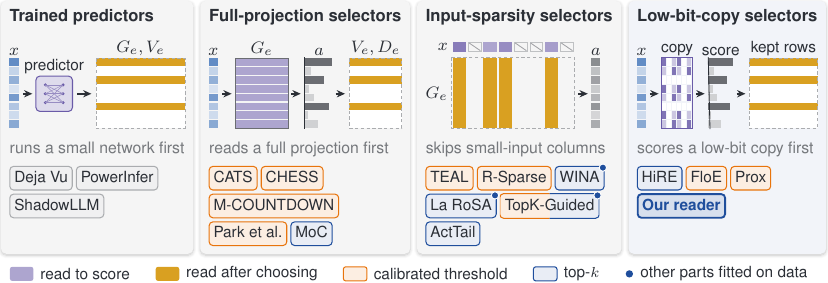}
\caption{\textbf{Prior channel selectors, grouped by what they read before choosing channels.}
Dashed rows or columns are skipped; chips show each keep rule, and a dot marks budgets or rotations fitted on data.
HiRE and Mixture-of-Channels (MoC) train the model for top-$k$; TopK-Guided bounds a threshold by a top-$k$ window.
Our reader (bold) fits nothing to the model.}\label{fig:related-map}
\end{figure}
\paragraph{Four families of channel selectors.} Prior selectors differ in what they read before choosing channels (Figure~\ref{fig:related-map}).
\emph{Trained predictors} use an auxiliary network \citep{liu2023dejavu,song2024powerinfer,akhauri2024shadowllm}, and SparseInfer uses sign bits without training \citep{shin2024sparseinfer}.
\emph{Full-projection selectors} compute a whole gate or up projection first, so they read at least a third of each routed expert \citep{lee2024cats,he2024chess,park2026intraexpert}, as does M-COUNTDOWN \citep{cheon2025countdown}.
\emph{Input-sparsity selectors} skip the weight columns that multiply small inputs \citep{liu2025teal,zhang2025rsparse,liu2025larosa,chen2025wina,acttail2026,wisparse2026}.
\emph{Low-bit-copy selectors}, including our reader, score channels with a compressed copy of the weights and then load the kept rows \citep{samaga2024hire,zhou2025floe,liu2026prox}; FloE computes its whole up projection from an INT2 copy.

\paragraph{Calibration and prior keep-rule comparisons.} Most selectors for frozen models calibrate on sample data, while HiRE and
Mixture-of-Channels \citep{wu2025moc} train the model for top-$k$. Six earlier studies compare the
two keep rules, and they disagree (Appendix Table~\ref{tab:keep-rule-prior}). La RoSA \citep{liu2025larosa}, Sparser-MoE
\citep{anonymous2026sparsermoe}, and batch-wise pruning \citep{kim2026batchwise} favor top-$k$. The
Sparsing Law \citep{luo2025sparsing} and Prox \citep{liu2026prox} favor thresholds, Prox by 1.8 points on
dense Qwen3 models. TopK-Guided \citep{agarwalla2026topkguided} finds a hybrid best. However, five of the six
studies test dense models, mostly on short tasks, and two change the score along with the keep rule.
Several works note that reads under a calibrated threshold drift at test time \citep{liu2025larosa,acttail2026,anonymous2026recomp}, yet none of the six runs top-$k$ at the reads a threshold actually used.
Prox's result does not carry over to OLMoE: on its likelihood tasks at 13\% and 18\% budgets, its two versions tie in 7 of 10 comparisons, and top-$k$ leads in 2 (Appendix~\ref{app:prox-olmoe}).

\paragraph{Scope.} Our conclusions concern calibrated thresholds on the channels of frozen MoE experts.
Task-dependent sparsity also motivates data-free selection \citep{szatkowski2026universal}, and SparseDecoding
finds that calibration and decoding activations differ \citep{wang2026sparsedecoding}. Thresholds can work elsewhere, as in sparse attention \citep{anonymous2026attnmass} or trained MoE routers
\citep{anonymous2026qualmoe}. Appendix~\ref{app:related} covers recovery, batching, and other levels of sparsity.

\section{Channel selection needs no calibration data}
\label{sec:findings}
\begin{figure}[!t]
\centering
\includegraphics[width=\linewidth]{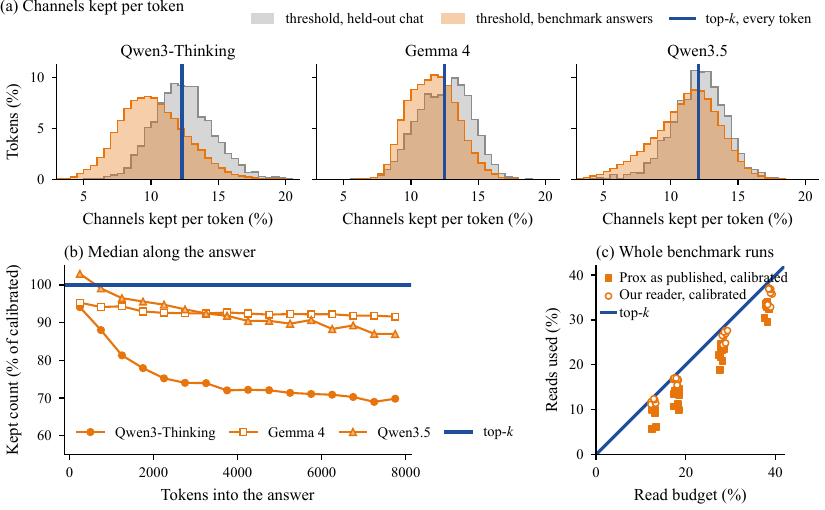}
\caption{\textbf{Under a calibrated threshold, the number of kept channels changes from token to token, and selectors read less than their budget on benchmarks.}
(a)~Share of channels each token keeps under an exact-score threshold calibrated on UltraChat for about 12\%, on held-out chat and on the model's answers to GPQA-Diamond \citep{rein2024gpqa}, IFEval
\citep{zhou2023ifeval}, and MMLU-Pro \citep{wang2024mmlupro}; top-$k$ keeps the same share on every token (line).
(b)~Median kept count per 500-token window, relative to the calibrated count (about
94{,}000 answer tokens per model). (c)~In all 72 benchmark runs, Prox and our reader with calibrated
thresholds read below their budget (the diagonal).}
\label{fig:drift}
\end{figure}
Selectors calibrate a threshold for two reasons: to read a known budget, and to gain accuracy at that budget by giving each token the channels it needs.

\subsection{A calibrated threshold does not fix how much a selector reads}
\label{sec:drift}
Channel scores change across tokens, tasks, and answer length, so a different number of channels passes a fixed cutoff $\tau$ each time, while top-$k$ always keeps $k$.

\paragraph{Reads vary across tokens and fall during generation.} Figures~\ref{fig:drift}a--b isolate the keep rule from the errors of an approximate score: channels are scored by their exact contribution to the output, and $\tau$ is calibrated on UltraChat to keep about 12\% (12.3\% on Qwen3-Thinking).
On held-out chat, the mean kept count matches the target, consistent with Prox's small drift on five short tasks \citep{liu2026prox}.
Yet even on chat, the middle 90\% of Qwen3-Thinking tokens keep 9.0 to 16.2\% of channels.
On benchmark answers, scores are smaller and the median falls to 10.0\% (Figure~\ref{fig:drift}a, orange).
Counts also fall as the answer grows (Figure~\ref{fig:drift}b): around 6{,}000 tokens, Qwen3-Thinking keeps 71\% of its calibrated count, Gemma~4 92\%, Qwen3.5 88\%.

\paragraph{Calibrated budgets are missed at deployment.} Selectors with calibrated thresholds read less than their budget in all 72 benchmark runs we made: Prox and our reader, each at targets of about 13\%, 18\%, 28\%, and 38\% (Figure~\ref{fig:drift}c; Appendix~\ref{app:drift}).
Prox with its threshold calibrated for 18\% reads 12.9\% on average and as little as 9.9\%.
About 1.7 percentage points of this gap are built into Prox's published rule for splitting the budget between inputs and channels, once reads are counted in sectors (Appendix~\ref{app:setup}).
Our reader with calibrated thresholds hits its target on calibration data but reads 0.4 to 5.8 percentage points less on benchmarks.
Calibrating on the models' own reasoning does not stop this drift (Appendix~\ref{app:own-generations}).

\paragraph{Kept channels follow the earlier answer, not token need.} Why do counts fall?
We take a 256-token window from position 6{,}000 of a Qwen3-Thinking GPQA-Diamond answer and place it directly after the question.
Under a Prox threshold calibrated for 13\%, the same tokens then keep 57\% more channels, close to the count at the start of the answer.
Removing earlier context raises the count even for the same tokens.
This fall barely helps: keeping each layer's count the same early and late in the answer, at equal total reads, changes KL divergence by at most $1.0\pm0.5\%$ on all three models.
Nor do harder tokens get more channels: kept counts correlate with the full model's next-token loss at $-0.29$ to $+0.14$ (Spearman).
Does the remaining variation from token to token help?
Giving every token in a layer the same count, at equal reads, lowers KL divergence on reasoning by $3.3\pm1.0\%$ on Qwen3-Thinking and $12.5\pm1.5\%$ on Gemma~4, so there the variation hurts; on Qwen3.5 it raises KL by $10.5\pm1.1\%$, so there it helps (Appendix~\ref{app:allocation}).

\subsection{Calibration gives no consistent accuracy gain over \texorpdfstring{top-$k$}{top-k}}
\label{sec:keep-rule}
We compare the threshold and top-$k$ versions of existing selectors (Section~\ref{sec:prelim}) at the budget each threshold was calibrated for.
Prox also selects inputs first; Appendix~\ref{app:keep-rule} gives the input rules.

\begin{figure}[!htb]
\centering
\includegraphics[width=\linewidth]{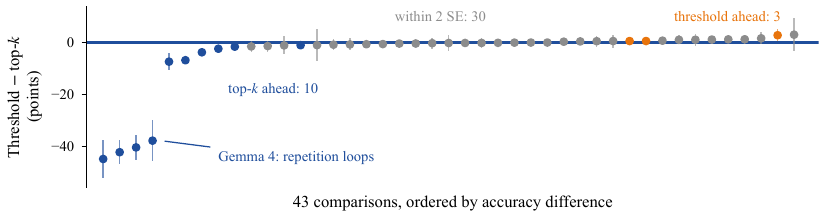}
\caption{\textbf{On the same score, a selector's top-$k$ version matches or beats its threshold version in 40 of 43 comparisons and avoids Gemma~4's collapse.}
Each dot is the accuracy of a selector's threshold version minus that of its top-$k$ version, at the budget the threshold was calibrated for, with $\pm2$ paired SE.
Gray marks ties, blue top-$k$ leads (the four on Gemma~4 by 38 to 45 points), and orange threshold leads (at most 2.8 points).
See Appendix Figure~\ref{fig:ranked-realized} for the same comparisons with each top-$k$ version run at the reads its threshold version actually used.}
\label{fig:ranked}
\end{figure}

\paragraph{Top-$k$ matches or beats thresholds across models and scores.} We make 43 paired comparisons at 13\% and 18\% budgets, on generation tasks on four MoE models and likelihood tasks on OLMoE, with three channel scores: Prox, TEAL, and Prox scaled to output units (Appendix~\ref{app:keep-rule}).
The top-$k$ version ties the threshold version in 30 (within two paired standard errors), leads in 10, and trails in 3, each time by at most 2.8 points (Figure~\ref{fig:ranked}).
We exclude two TEAL comparisons where both versions collapse.
Pooled per model, no model favors the threshold version (Appendix Table~\ref{tab:keep-pooled}).
On the three post-trained models alone, the top-$k$ version leads in 5 of 11 comparisons and the threshold version in none.

\paragraph{Robustness at realized reads.} We also run each top-$k$ version at the reads its threshold version actually used, a check that favors the threshold version (Section~\ref{sec:prelim}).
The top-$k$ version still matches or beats it in 37 of 41 comparisons (Appendix~\ref{app:keep-rule}).
The largest lead of a threshold version, $+11.5\pm1.9$ points for Prox on Qwen3-Thinking IFEval, disappears at the calibrated budget, where Prox scores 85.4 with top-$k$ and 84.3 with its threshold.
At similar reads, controls show roughly equal gains from per-layer budgets and from per-token counts (Appendix~\ref{app:allocation}).
Per-token variation helps most on Qwen3.5 GPQA-Diamond, by $+7.1\pm2.7$ points.
But this gain comes at a budget the deployer did not choose: with its threshold calibrated for 13\%, Prox reads only 9.1\% there and scores 80.6, which ties the 79.9 it scores with top-$k$ at 13.5\% of reads.

\paragraph{Other scores and calibrated parts.} Our reader behaves the same way.
Replacing its top-$k$ with calibrated per-layer thresholds gives 34 ties and 2 comparisons where the top-$k$ version leads, both on Gemma~4, out of 36.
CATS \citep{lee2024cats}, which thresholds exact gate activations, ties in 16 of 18 comparisons at 53\% and 67\% of reads, and its threshold version leads in the other 2, both on Gemma~4 (Appendix~\ref{app:calibrated-reader}).
Some selectors also calibrate layer budgets (TEAL, WINA) or recovery terms (R-Sparse, GRAIL).
Calibrated versions of these parts lower KL divergence, calibrated layer budgets even on reasoning text unlike their calibration data, but improve no benchmark beyond two standard errors; a calibrated recovery map beats our weight-only map in 0 of 18 comparisons~(Appendix~\ref{app:calibrated-parts}).

\subsection{Long reasoning can collapse under calibrated thresholds}
\label{sec:collapse}

\paragraph{Channel thresholds trigger repetition loops on Gemma~4.} The four largest gaps in Figure~\ref{fig:ranked} share one cause: on Gemma~4 at the 13\% budget, the top-$k$ versions of Prox and its output-scaled variant beat their threshold versions by 38 to 45 points.
With a calibrated threshold, Gemma~4 repeats phrases until the 32{,}768-token limit.
Reading too little does not explain this.
Over four samples per question, Prox with its threshold calibrated for 18\% reads 13.3\% and scores 11.7.
With top-$k$ at nearly the same reads, 13.5\%, Prox scores 56.3 (Appendix Figure~\ref{fig:headline-grid-realized}), and 145 rather than 674 of 792 answers hit the limit.
Replacing only Prox's channel threshold with top-$k$ avoids the collapse; replacing only its input threshold does not (Appendix~\ref{app:gemma}).
Neither a 4\% floor on channels nor calibration on Gemma~4's own reasoning prevents it (Appendices~\ref{app:own-generations} and~\ref{app:gemma}).
Similar loops appear in other compressed models: pruning amplifies loops at stable perplexity \citep{lee2026focus}, 2-bit quantization often makes reasoning loop or exhaust its token budget \citep{alimaskina2026extremelowbit}, and batch-shared TEAL masks collapse reasoning \citep{kim2026batchwise}.

\paragraph{Collapse behaves like a cliff.} We suspect that once a sparse model starts a repetition, it rarely stops on its own.
If so, accuracy drops sharply, not gradually, once the error sparsity adds to each token passes a level that depends on the model, and a less accurate selector needs more reads to stay below it.
On Gemma~4 IFEval, where the full model scores 93.5, WINA as published scores 40.9 at 20\% of reads, and 53\% of its answers stop at the length limit.
Prox's score with top-$k$ reaches 57.1 at 13.5\% of reads, and our reader 88.0 at 12.8\%.
Prox still collapses when each layer's count is fixed to its average under the threshold (Appendix~\ref{app:allocation}).
With a calibrated threshold, a deployer therefore cannot see this level coming: the selector's reads are not known in advance (Section~\ref{sec:drift}), and even at equal reads its accuracy can fall far below that of its top-$k$ version (11.7 against 56.3 above).

\section{A reader built from the weights alone}
\label{sec:method}

With top-$k$ as the keep rule, a selector's quality depends on two things: how well it scores channels before reading them, and how it handles the channels it skips. We build our reader around both.

\paragraph{Scoring matters more than the keep rule.} Switching Prox from its calibrated threshold to top-$k$ changes next-token KL divergence by only $-17\%$ to $-3\%$ on held-out OLMoE documents and by $-6\%$ to $+24\%$ on chat.
Changing the score moves KL much more.
On the same 4-bit weight copy, scaling Prox's score to output units, with down-projection norms and router weights, divides KL by 1.3 to 1.5.
On OLMoE development data at 17.8\% of reads, with no recovery, selectors that score with an INT4 copy of the gate projection or with TEAL's input magnitudes reach 2.8 and 5.1 times the KL divergence of our reader's full-sketch screen (Appendix Table~\ref{tab:design}).
Across four scores and four budgets on OLMoE, KL divergence falls as the kept channels carry a larger share of the routed output's energy (Spearman $-0.96$; Appendix~\ref{app:components}).

\begin{figure}[!htb]
\centering
\includegraphics[width=0.88\linewidth]{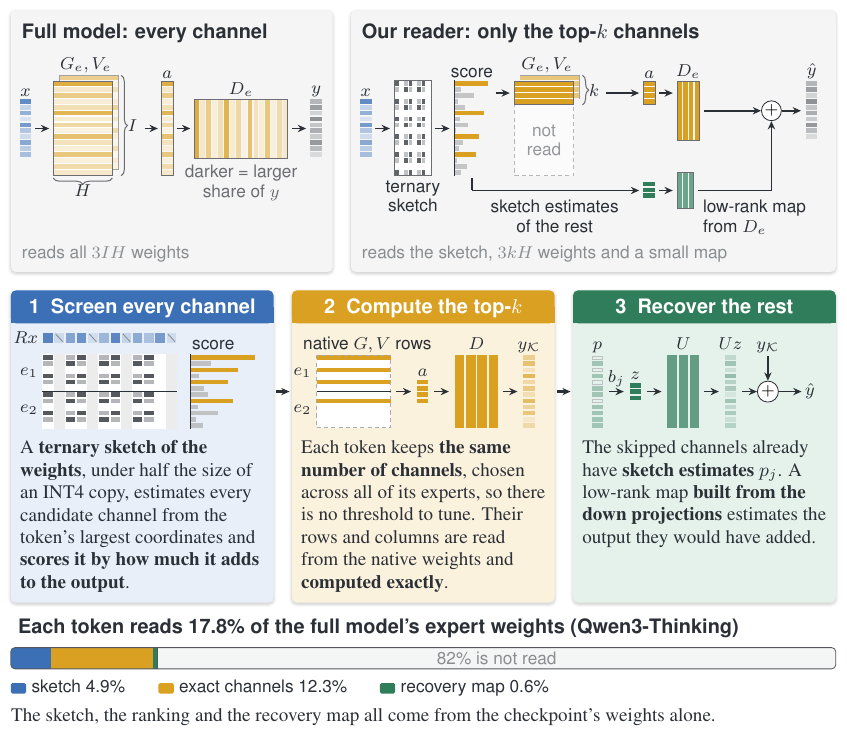}
\caption{\textbf{Our reader screens candidate channels with a ternary sketch, computes the top-$k$
channels exactly, and recovers the rest using only the full model's weights.}
Top: the full model reads every channel of a routed expert; our reader reads a ternary sketch, the $k$ kept
channels, and a low-rank recovery map. Middle: the three stages. Bottom: reads at 17.8\% on Qwen3-Thinking.}
\label{fig:method}
\end{figure}

Our reader therefore puts its effort into the score and into recovering skipped channels.
It is built once from the model's weights, with nothing fitted to the target model.
Three settings were chosen once on OLMoE development data and reused on every model: 128-coordinate rotation blocks, 80-coordinate partial screens, and rank-32 recovery (Appendix~\ref{app:components}).
For each token and layer, it screens all candidate channels, computes the top-$k$ exactly, and recovers the rest (Figure~\ref{fig:method}).

\subsection{Screening channels with a rotated ternary sketch}
\label{sec:screen}

The screen estimates every candidate channel's gate and up activations from a ternary sketch of the weights, built once at load time.
Weights and hidden states contain outliers, which make low-bit codes inaccurate.
We therefore rotate each gate row $g_j$ and up row $v_j$ with a randomized Hadamard transform $R$ \citep{ashkboos2024quarot,tseng2024quipsharp}, applied in blocks of 128 coordinates, which spreads large values across each block.
Because $R$ is orthogonal, the rotation leaves activations unchanged: $g_j^\top x=(Rg_j)^\top(Rx)$.
Each block of $Rg_j$ and $Rv_j$ is stored as ternary codes $\{-1,0,1\}$ \citep{li2016ternary}, together with one FP16 scale and the block's mean-squared quantization error $\rho_b$.
This shrinks a block from 256 bytes in BF16 to about 30 bytes, so the sketch adds about 7\% to resident memory.
At decode time, the reader rotates $x$ once and multiplies it by the sketch, giving estimates $\hat g_j\approx g_j^\top x$ and $\hat v_j\approx v_j^\top x$.

At our two lowest budgets, 12.8\% and 17.8\% of reads, reading the full sketch would take 44 to 61\% of all reads on OLMoE. At these two budgets the reader runs a \emph{partial screen}: in each block, it reads ternary codes only at the 80 coordinates where $|Rx|$ is largest.
This buys 20 to 50\% more exact channels.

\subsection{Output-scaled scoring and exact \texorpdfstring{top-$k$}{top-k} computation}
\label{sec:select}

The reader scores each candidate channel by the size of what it would add to the MoE output.
Channel $j$ adds $w_j\,a_j\,d_j$, where $w_j$ is the router weight of channel $j$'s expert, $a_j=\phi(g_j^\top x)(v_j^\top x)$ its activation, and $d_j$ its down-projection column (Figure~\ref{fig:prelim}a).
The size of this contribution is $w_j\,\|d_j\|\,|a_j|$, and replacing $a_j$ with its sketch estimate gives the score
\begin{equation}
  s_j=w_j\,\|d_j\|\,\bigl|\phi(\hat g_j)\,\hat v_j\bigr| ,
  \label{eq:score}
\end{equation}
where the norms $\|d_j\|$ are computed once at load time.
Because the score is in output units, channels from different experts are directly comparable.
The reader therefore picks the $k$ best channels across all of the token's experts, and one expert may receive many channels while another receives none.
From 28.7\% of reads upward, the reader reads the full sketch and also uses its block errors $\rho_b$ to rank channels by the expected squared output error left after recovery if skipped (Appendix~\ref{app:components}).

For the $k$ kept channels, the reader loads the full weights and computes the exact activations $a_j$.
To make each kept channel's down weights contiguous, we transpose every down projection $D_e$ once at load time, so that column $j$ becomes a row.
Prefill and decode both use this one layout.

\subsection{Weight-only low-rank recovery of skipped channels}
\label{sec:recover}

Each skipped channel contributes little, but thousands of them per layer together matter.
The screen has already estimated the activation $p_j=\phi(\hat g_j)\hat v_j$ of every skipped channel $j\in\mathcal{S}$, so the reader can reuse these estimates without rereading any gate or up weights.
It still needs each skipped channel's down column $d_j$ to turn the estimate into output.
To avoid loading these full-width columns, we approximate them with a small map built once before deployment.
We stack the down columns of all experts in a layer, compute their top 32 left singular vectors $U\in\mathbb{R}^{H\times 32}$, and store each channel's coefficients $b_j=U^\top d_j$.
With kept channels $\mathcal{K}$, the reader then outputs
\begin{equation}
  \hat y=\sum_{j\in\mathcal{K}}w_j\,d_j\,a_j\;+\;U\sum_{j\in\mathcal{S}}w_j\,b_j\,p_j .
  \label{eq:output}
\end{equation}
The first sum is exact; the second adds the skipped channels' estimated output through the rank-32 map. The map adds about 0.5\% to resident memory and 0.5 to 0.8 percentage points of reads.

\paragraph{Each component earns its reads.} On OLMoE at equal reads, spending 0.6 to 0.7 percentage points on recovery lowers KL divergence more than spending them on more exact channels.
The partial screen lowers KL relative to the full screen, from 0.488 to 0.310 at 13.5\% of reads and from 0.245 to 0.187 at 18.5\%.
The same components help benchmark accuracy on Gemma~4.
With the full screen at 18\% of reads, recovery raises IFEval accuracy by $11.6\pm1.6$ points and reduces repetition loops.
At 13\%, the partial screen beats the full screen by $13.7\pm1.3$ points on MMLU-Pro.

\section{Accuracy and speed on post-trained MoE models}
\label{sec:results}

\subsection{Accuracy}
\label{sec:accuracy}

\paragraph{Experimental setup and baselines.} We evaluate our reader in vLLM \citep{kwon2023vllm} on three post-trained MoE models, Qwen3-Thinking, Gemma~4, and Qwen3.5, and three benchmarks, GPQA-Diamond, IFEval, and MMLU-Pro (Appendix~\ref{app:setup}).
The full model processes each prompt, and the selector runs on every decoded token.
We compare against three published selectors.
Prox \citep{liu2026prox} uses calibrated thresholds, and we also run its score with top-$k$.
TEAL \citep{liu2025teal} uses calibrated thresholds with greedy per-layer budgets.
WINA \citep{chen2025wina} as published keeps the top-$k$ with calibrated per-projection budgets, and we also run its score with calibrated thresholds.
Every calibrated part is fitted on the same UltraChat assistant turns \citep{ding2023ultrachat}.
We report accuracy as a percentage of the full model's accuracy, over nine model-benchmark pairs (Appendix Table~\ref{tab:headline}), and count a difference only beyond two question-clustered paired standard errors.

\begin{table}[!htb]
\centering
\caption{\textbf{Our reader is the most accurate tested selector at the two lowest budgets.}
Mean and worst accuracy over nine model-benchmark pairs, as a percentage of the full model's accuracy.
Selectors with calibrated thresholds list the budget they were calibrated for (Target) and the reads they
used (Realized); WINA as published uses top-$k$ and meets its budget to the precision shown.}
\label{tab:accuracy}
\small
\setlength{\tabcolsep}{4pt}
\begin{tabular}{lrrrrrrrr}
\toprule
& \multicolumn{4}{c}{About 13\% of reads} & \multicolumn{4}{c}{About 18\% of reads} \\
\cmidrule(lr){2-5}\cmidrule(lr){6-9}
Selector & Target & Realized & Mean & Worst & Target & Realized & Mean & Worst \\
\midrule
Our reader & -- & 12.8 & \textbf{94.5} & \textbf{86.9} & -- & 17.8 & \textbf{98.1} & \textbf{95.0} \\
Prox score, top-$k$ & -- & 13.5 & 87.0 & 61.1 & -- & 18.5 & 96.1 & 90.4 \\
Prox as published, calibrated & 13 & 8.5 & 51.5 & 0.0 & 18 & 12.9 & 68.5 & 12.1 \\
WINA as published, top-$k$ & -- & 15 & 2.4 & 0.0 & -- & 20 & 28.4 & 1.1 \\
WINA score, calibrated & 15 & 7.8 & 0.2 & 0.0 & 20 & 11.2 & 5.0 & 0.0 \\
TEAL as published, calibrated & 15 & 8.9 & 0.2 & 0.0 & 20 & 12.6 & 9.9 & 0.0 \\
\bottomrule
\end{tabular}
\end{table}

\paragraph{Accuracy at heavy sparsity.} Our reader has the highest mean accuracy at both low budgets, and its weakest pair beats every other selector's weakest pair (Table~\ref{tab:accuracy}; Appendix Figure~\ref{fig:headline-grid} shows every pair).

Prox, with its threshold calibrated for 18\%, reads 12.9\%, about as much as our reader at 12.8\%.
Yet it reaches only 68.5\% of full-model accuracy on average, against 94.5\% for our reader.
Most of this gap comes from Gemma~4, where its threshold version falls into the repetition loops of Section~\ref{sec:collapse}; on the two Qwen models, Prox is about as accurate as our reader at similar reads.
Top-$k$ cannot rescue a weak score: WINA as published uses top-$k$ and reaches only 28.4\% of full-model accuracy at 20\% of reads.
From 28\% of reads upward, our reader and Prox, with either keep rule, stay within 10\% of the full model.

\subsection{Speed}
\label{sec:speed}

\begin{figure}[!htb]
\centering
\includegraphics[width=\linewidth]{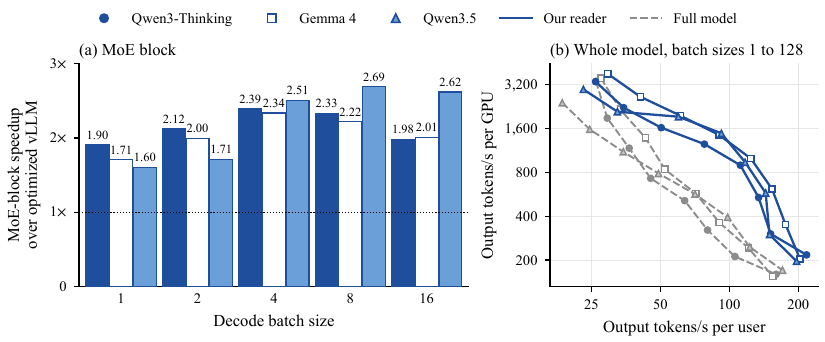}
\caption{\textbf{Our reader accelerates MoE blocks by 1.60 to 2.69$\times$ over optimized vLLM across three post-trained models.}
At 17.8\% of routed-expert reads, median of three processes. (a)~MoE-block decode speedup; both arms share the unmodified router, and Gemma~4's dense MLP and Qwen3.5's shared expert are timed inside the block. (b)~Whole model, batch sizes 1 to 128: right is faster per user, up is more output per GPU.}
\label{fig:speed}
\end{figure}

\paragraph{Our reader accelerates MoE blocks by up to 2.69$\times$ in optimized vLLM.} Our Triton kernels select exactly the channels of the reader evaluated above, at 17.8\% of reads on one RTX PRO 6000 Blackwell GPU. We compare against unmodified vLLM in its fastest configuration, default or tuned, at each batch size, with the same router (Figure~\ref{fig:speed}).
On Qwen3-Thinking, Gemma~4, and Qwen3.5, MoE blocks run 1.90$\times$, 1.71$\times$, and 1.60$\times$ faster at batch one, and 2.39$\times$, 2.34$\times$, and 2.51$\times$ at four.

The whole model gains less, because attention runs as before.
Whole models decode 1.15$\times$ to 1.36$\times$ faster at batch one and up to 1.75$\times$, 1.74$\times$, and 1.88$\times$ at batch eight or sixteen, in the same model order, and 1.08$\times$ to 1.54$\times$ with 8k to 32k tokens of context.
At 100 tokens per second per user, one GPU serves 3.3 to 4.3 times as many tokens (Figure~\ref{fig:speed}b; Appendix~\ref{app:speed}).

At batch one, the speedup is smaller than the saved reads suggest.
On Qwen3-Thinking, the MoE block moves only 18.9\% of the full model's memory traffic, yet it runs 1.90$\times$ faster, not about 5$\times$.
Part of each step is a fixed cost, such as launching kernels, that fewer reads do not shrink.
Larger batches share these fixed costs, so the block speedup grows up to batch four or eight (Table~\ref{tab:speed}).

In dense models, the channels a batch needs together grow with batch size and approach the dense set in deeper layers, so channel selection helps mainly at small batches \citep{shrestha2025polar}.
MoE models with many small experts push this limit to larger batches: at batch 16, each expert serves about one token or fewer on average.
Our reader still decodes 1.72$\times$ to 1.88$\times$ faster as a whole model at this batch size, and at batch 128 its speedup falls to 1.00$\times$ to 1.24$\times$.
\citet{park2026intraexpert} also find that their MoE-layer speedup peaks at batch 16 to 128.
On our GPU, their published kernels beat our reader's MoE block only at batch 64 and 128.
As a whole model, our reader is faster than their kernels at every batch size up to 16.
Their kernels do not fit in GPU memory at batch 32 and larger (Appendix~\ref{app:speed}).

\section{Limitations and conclusion}
\label{sec:conclusion}
\paragraph{Limitations.} Our kernels turn only part of the saved reads into speed (Section~\ref{sec:speed}), and they were tuned for one budget on one GPU type (Appendix~\ref{app:speed}).
Selecting channels for a whole batch at once, so the gain lasts to larger batches, is future work.
The share of coordinates the partial screen reads was chosen on OLMoE and checked only on Gemma~4 (Table~\ref{tab:design}).
Our reader changes only routed experts in decoding.
Its sketch and recovery map add about 7\% and 0.5\% of a BF16 checkpoint, a larger share of a quantized one (Table~\ref{tab:models}).
We tested models of up to 35B parameters.

\paragraph{What our results suggest.} We suspect two patterns hold beyond our models.
First, short likelihood evaluations hide failures of long generation.
On likelihood tasks, the two versions of a selector differ little and in both directions.
On generation tasks at a 13\% or 18\% budget, no threshold version beats its top-$k$ version, and Gemma~4 falls into repetition loops under thresholds.
REAP \citep{lasby2025reap} saw the same contrast when merging experts, and pruning can add loops at largely unchanged perplexity \citep{lee2026focus}.
Second, lower KL divergence need not mean higher accuracy: calibrated layer budgets lower KL divergence even on reasoning text unlike their calibration data, yet neither they nor calibrated recovery maps raise a benchmark score.

\paragraph{Conclusion.} Channel selection for MoE decoding needs no calibration data.
With a calibrated threshold, a selector's reads follow the input rather than the budget.
At the calibrated budget, its threshold version gains no consistent accuracy over its top-$k$ version on the same score.
Built from the weights alone, our reader retains 98.1\% of full-model accuracy at 17.8\% of routed-expert reads on three models and runs MoE blocks up to 2.69$\times$ faster and whole models up to 1.88$\times$ faster than optimized vLLM.

\subsubsection*{AI use statement}
We used LLM-based coding assistants to write the experiment code and kernels, run experiments, and write the
scripts that summarize results, and to help draft and edit the text. The authors designed the study, checked
every reported number against the experiment outputs, revised the manuscript, and take full responsibility
for its content.

\subsubsection*{Reproducibility statement}
All models, benchmarks and calibration data we use are publicly available. Sections~\ref{sec:prelim}
and~\ref{sec:method} define the selectors and our reader. Appendix~\ref{app:setup} gives the models,
benchmarks, generation settings, statistics, calibration protocols, baseline adaptations and hardware,
and Appendix~\ref{app:accounting} defines how reads are counted. Our reader, its Triton kernels,
the vLLM patches, the baselines, the evaluation scripts and the scripts that compute the paper's
numbers from the evaluation results are in
\ifarxiv\url{https://github.com/AidenGeunGeun/calibration-free-moe}\else
an anonymized repository: \url{https://anonymous.4open.science/r/calibration-free-moe-4727/}\fi.

\bibliography{refs}
\bibliographystyle{iclr2027_conference}

\clearpage
\appendix
\etocdepthtag.toc{appendix}
\etocsettagdepth{main}{none}
\etocsettagdepth{appendix}{subsection}
\etocsettocstyle{\section*{Appendix contents}}{}
\tableofcontents
\clearpage

\section{Extended related work}
\label{app:related}

\begin{table}[tb]
\centering
\caption{\textbf{Prior keep-rule comparisons reach opposite conclusions, focus mostly on short tasks on dense
models, and none runs top-$k$ at the reads a threshold actually uses.}
Unlike most prior comparisons, ours holds the channel score fixed (\emph{Same score}), tests long
chain-of-thought reasoning on MoE models (\emph{Regime}), and runs top-$k$ both at the target budget and at
realized reads (\emph{Matched}).}
\label{tab:keep-rule-prior}
\small
\setlength{\tabcolsep}{2.5pt}
\begin{tabular*}{\linewidth}{@{\extracolsep{\fill}}lllcll@{}}
\toprule
Study & Regime & Models & \begin{tabular}[b]{@{}c@{}}Same\\score\end{tabular} & Matched & Ahead \\
\midrule
La RoSA \citeyearpar{liu2025larosa} & dense, PPL & LLaMA2-7B & yes & target & top-$k$ (PPL) \\
Batch-wise pruning \citeyearpar{kim2026batchwise} & dense, CoT & 2 distilled 7--8B & yes & target & top-$k$ (batch 4) \\
Sparsing Law \citeyearpar{luo2025sparsing} & dense, short tasks & 0.1B, 0.8B & no & target & threshold \\
Prox \citeyearpar{liu2026prox} & dense, short tasks & Qwen3-8B, 14B & yes & target & threshold (+1.8 pts) \\
TopK-Guided \citeyearpar{agarwalla2026topkguided} & dense, short tasks & Llama-2/3, 7--8B & no & target & mixed; hybrid best \\
Sparser-MoE \citeyearpar{anonymous2026sparsermoe} & MoE, short tasks & OLMoE & unclear & target & top-$k$ ($<$1 pt) \\
\midrule
This paper (\S\ref{sec:keep-rule}) & MoE, short + CoT & 4 MoE models & yes & both & tie; top-$k$ on Gemma~4 \\
\bottomrule
\end{tabular*}
\end{table}

\paragraph{Prior comparisons of calibrated thresholds and top-$k$.} Earlier studies disagree on whether a
calibrated threshold or top-$k$ selects channels better (Table~\ref{tab:keep-rule-prior}). La RoSA,
Sparser-MoE, and batch-wise adaptive pruning find top-$k$ ahead; batch-wise adaptive pruning is the only prior study in the
table that tests long chain-of-thought reasoning. The Sparsing Law and Prox find calibrated thresholds
ahead. TopK-Guided finds top-$k$ (WINA) ahead of thresholded TEAL on Llama-2-7B but not consistently on Llama-3-8B, and its best rule is a
hybrid: a calibrated threshold whose per-token count stays within $\pm\delta$ of $k$, combined with calibrated
block budgets. In our MoE experiments, calibrated layer budgets give no held-out benchmark gain
(Appendix~\ref{app:calibrated-parts}), and a per-token floor on the count does not prevent Gemma~4's collapse
(Appendix~\ref{app:gemma}). We do not test a two-sided $\pm\delta$ band.

\paragraph{Approximating skipped channels.} Instead of simply dropping the channels they skip, several
methods approximate what those channels would have added. R-Sparse adds low-rank weight factors from an SVD
and chooses how to split the budget between them and sparse inputs on calibration data. GRAIL \citep{tang2026grail} fits linear
compensation maps on calibration data to approximate omitted activations. ReComp \citep{anonymous2026recomp}
distills a calibrated low-rank estimate of the skipped MLP output. Other methods send medium-importance
columns through a compressed copy, as in SpAx \citep{kim2026spax}, or through a pruned or low-precision path,
as in DAWS \citep{sun2026daws} and SharQ \citep{meng2026sharq}. Our recovery map needs no calibration data. It reuses each skipped channel's
activation estimate from the ternary screen and passes it through the top singular vectors of the down
projections.

\paragraph{Speed under batching.} In batched decoding, the GPU must load every channel that any token in the
batch selects, so per-token selection saves fewer reads as the batch grows \citep{shrestha2025polar}. SERE
\citep{wu2026sere} addresses this by shrinking the union of routed experts across a batch. Weak dense baselines
can also inflate reported speedups \citep{lee2026bytecross}. We therefore charge reads in hardware sectors,
check them with a GPU profiler, time against optimized vLLM, and report speedups at batch sizes one to
sixteen.

\paragraph{Other levels of sparsity.} Other methods act at a different granularity and can be combined with
our reader. Expert-skipping and expert-merging methods drop,
prune, or fold whole experts
\citep{lu2024naee,modes2025,chittyvenkata2025lexi,lasby2025reap,halving2026,xu2026ace,wu2026exfold,abbasi2026base,kim2026lda}.
Static compression methods prune or quantize weights ahead of time
\citep{frantar2023sparsegpt,frantar2023gptq,sun2024wanda,an2024flap,liu2025spinquant,behera2026lila}.
Finally, prior work has studied per-token channel selection \citep{samaga2024hire}, adaptive pruning with little or no
calibration data \citep{dong2024griffin,tyagi2026dart,kim2026batchwise}, and sparsity inside MoE experts
\citep{zhou2025floe,park2026intraexpert}, including top-$k$ neuron selection trained into experts during
pretraining \citep{cheng2025mone}.

\section{Experimental setup}
\label{app:setup}

\begin{table}[tb]
\centering
\caption{\textbf{Evaluated MoE models and the resident memory our reader adds.} Each row gives the
number of routed experts and how many each token uses, the channels per expert and the hidden width, the
number of MoE layers, any module that always runs in full, and the resident memory of our reader's ternary
sketch and recovery map as a percentage of the BF16 checkpoint. OLMoE's sketch size is computed analytically.}
\label{tab:models}
\small
\begin{tabular}{lcccccccc}
\toprule
 & \multicolumn{2}{c}{Routed experts} & \multicolumn{2}{c}{Width} & MoE & Always & \multicolumn{2}{c}{Added memory (\%)} \\
\cmidrule(lr){2-3}\cmidrule(lr){4-5}\cmidrule(lr){8-9}
Model & Total & Active & Expert & Hidden & layers & full & Sketch & Recovery \\
\midrule
Qwen3-Thinking & 128 & 8 & 768 & 2048 & 48 & none & 7.6 & 0.50 \\
Gemma~4 & 128 & 8 & 704 & 2816 & 30 & dense MLP & 7.0 & 0.35 \\
Qwen3.5 & 256 & 8 & 512 & 2048 & 40 & shared expert & 7.0 & 0.47 \\
OLMoE & 64 & 8 & 1024 & 2048 & 16 & none & 7.3 & 0.50 \\
\bottomrule
\end{tabular}

\end{table}

\begin{table}[tb]
\centering
\caption{\textbf{Evaluation protocol on the post-trained models.} Each model runs in thinking mode with its model card's sampling parameters and random seed 20260923.
GPQA-Diamond reports accuracy over four samples per question, IFEval reports prompt-level strict accuracy, and MMLU-Pro reports accuracy on a fixed 1{,}000-question subset.}
\label{tab:protocol}
\small
\begin{tabular}{llcccccr}
\toprule
 & & & & \multicolumn{3}{c}{Sampling} \\
\cmidrule(lr){5-7}
Model & Benchmark & Items & Samples & Temperature & Top-$p$ & Top-$k$ & Max tokens \\
\midrule
Qwen3-Thinking & GPQA-D & 198 & 4 & 0.6 & 0.95 & 20 & 32{,}768 \\
 & IFEval & 541 & 1 & 0.6 & 0.95 & 20 & 32{,}768 \\
 & MMLU-Pro & 1{,}000 & 1 & 0.6 & 0.95 & 20 & 32{,}768 \\
\addlinespace
Gemma~4 & GPQA-D & 198 & 4 & 1.0 & 0.95 & 64 & 32{,}768 \\
 & IFEval & 541 & 1 & 1.0 & 0.95 & 64 & 32{,}768 \\
 & MMLU-Pro & 1{,}000 & 1 & 1.0 & 0.95 & 64 & 32{,}768 \\
\addlinespace
Qwen3.5$^{\dagger}$ & GPQA-D & 198 & 4 & 1.0 & 0.95 & 20 & 81{,}920 \\
 & IFEval & 541 & 1 & 1.0 & 0.95 & 20 & 32{,}768 \\
 & MMLU-Pro & 1{,}000 & 1 & 1.0 & 0.95 & 20 & 32{,}768 \\
\bottomrule
\multicolumn{8}{l}{$^{\dagger}$Presence penalty 1.5.} \\
\end{tabular}

\end{table}

\paragraph{Models and runtime.} We evaluate Qwen3-30B-A3B-Thinking-2507 (Qwen3-Thinking), Gemma-4-26B-A4B-it
(Gemma~4), and Qwen3.5-35B-A3B (Qwen3.5) in their official BF16 releases in vLLM 0.29.0. For every prompt, the full model processes the
prompt and generates the first output token, and the channel selector runs on every later decoded token.
Shared experts and parallel dense MLPs always run in full and are not counted in routed-expert reads.

\paragraph{Benchmarks.} GPQA-Diamond uses all 198 questions with four sampled answers per question, in each
model card's prompt format and with its sampling parameters. IFEval uses all 541 prompts and reports
prompt-level strict accuracy. MMLU-Pro uses a fixed stratified subset of 1{,}000 questions. Answers may be up
to 32{,}768 tokens long, except on Qwen3.5 GPQA-Diamond, where its model card recommends 81{,}920 tokens. The
full Qwen3-Thinking model also ran GPQA-Diamond with an 81{,}920-token cap, but none of its answers exceeds
23{,}500 tokens. On OLMoE, GSM8K \citep{cobbe2021gsm8k} (1{,}319 questions, eight-shot chain of thought) and BBH
\citep{suzgun2023bbh} (6{,}511 questions, three-shot chain of thought) use greedy decoding up to 512 tokens
under the task definitions of the OLMES evaluation suite \citep{gu2025olmes}.

\paragraph{Answers that hit the length limit.} On the post-trained models, an answer that reaches the token
limit without finishing counts as wrong, for every method. On IFEval, most such answers contain no final answer at all.
Of the 10{,}872 IFEval answers that hit the limit in our main runs, 10{,}649 never left their reasoning trace.
Of the other 223, 202 ended in repetition loops and the rest stopped mid-sentence. IFEval's format checkers
can still pass such fragments on rules such as lowercase-only or no commas, and crediting them would change
single model-benchmark results by up to 3.3 points. Counting them as wrong avoids crediting collapsed outputs.

\paragraph{Statistics.} Benchmark comparisons use paired standard errors clustered by question.
We count two methods as different when their paired difference exceeds two standard errors, without correcting for the number of comparisons.
A tie therefore means no difference was detected, not that the methods are equivalent.
Unless stated otherwise, $\pm$ denotes one standard error.
Next-token KL divergence is
$\mathrm{KL}(P\,\|\,\hat P)$, from the full model's next-token distribution $P$ to the sparse model's $\hat P$,
averaged over scored positions; top-1 agreement is the share of positions where both put the same token first.

\paragraph{Calibration on the post-trained models.} Every calibrated method is fitted on the same UltraChat
conversations \citep{ding2023ultrachat}. We split them with a fixed seed into 448 fitting, 64 tuning, and 128
development conversations; the tuning set only selects the checkpoint of the calibrated recovery map. Each
conversation is rendered with the model's chat template and contributes one 128-token window from an
assistant turn. Prox fits its thresholds on 20{,}480 token positions per layer, sampled from the full text of
30 fitting conversations. TEAL's greedy budget allocation and both WINA variants use 512 fitting states per
layer. UltraChat turns contain no reasoning traces, so these thresholds are fitted outside thinking mode and
evaluated in thinking mode. Appendix~\ref{app:own-generations} repeats the calibration on the models' own
reasoning instead.

\paragraph{How Prox splits its budget.} Prox splits each target budget between input and channel sparsity
with its published rule, which prices its 4-bit weight copy with Prox's latency model. Counted in sectors,
this rule plans 1.6 to 1.7 percentage points of reads below the target at budgets from 13 to 38\%, and 5.6
points below at 58 and 78\%. Qwen3.5's 13\% thresholds were fitted for a 12.5\% target. At test time, every
Prox run reads even less than this plan.

\paragraph{Calibration on OLMoE.} OLMoE is a base model, so its thresholds are fitted on Pile text
\citep{gao2020pile}. Prox uses the first 30 fitting documents (20{,}480 positions per layer), TEAL uses 177
documents (512 states per layer), and 40 further documents serve as tuning data. Our reader's constants and the
baselines' input fractions were chosen on a separate set of 192 development documents, and the held-out OLMoE
evaluation uses another 192. The development and held-out sets are disjoint from each other and from all fitting and tuning documents. Fitting Prox takes 3 to 19 minutes on a single
GPU.

\paragraph{Baselines.} Prox \citep{liu2026prox} was designed for dense models. We apply it directly to each
routed expert and store its 4-bit copy of the gate and up weights in the memory layout that needs the fewest
sector reads. Prox's top-$k$ version keeps each token's 30\% largest input coordinates (50\% from 28\% of reads
upward) and then its top-$k$ channels; these input fractions were also chosen on OLMoE development data. TEAL
\citep{liu2025teal} uses its greedy search to divide sparsity among the projections of each layer; we apply it to the gate, up, and
down projections of the routed experts. WINA's score with a calibrated threshold uses one threshold per projection.

WINA as published \citep{chen2025wina} scores each input coordinate by its magnitude times the matching column
norm of the weight matrix that consumes it (the gate projection in a gated MLP), keeps the top-$k$ per token, and fits a sparsity level for each projection with
TEAL's greedy search. We adapt this search to routed experts. It minimizes the reconstruction error of the
routed-expert output, uses empirical quantile thresholds measured on the fitting data, steps in increments of
0.01, and stops at the first sparsity level that reaches the target budget. Neither WINA variant applies
the orthogonal weight transform that WINA adopts to maximize its gains. CATS \citep{lee2024cats} computes the full gate
projection and keeps channels by exact gate magnitude with a calibrated threshold; we also run a per-token top-$k$ version.
Published HiRE trains the model with a group-sparse FFN; on frozen OLMoE we test only its screen, as a
HiRE-style control that screens channels with an INT4 copy of the gate projection under our top-$k$.

\paragraph{Hardware.} Benchmark runs used a single RTX PRO 6000 Blackwell GPU, two A6000 GPUs, or four A5000
or RTX 3090 GPUs. In the multi-GPU setups, attention is split by tensor parallelism and routed experts by expert
parallelism. Gemma~4 ran only on the first two setups. Before pooling results across these layouts, we ran a
consistency test fixed in advance. Every layout replayed 192 held-out chat documents of 128 tokens each for the
full model, our reader at 17.8\% of reads, and Prox calibrated for 18\%. For each sparse method, the 90\%
confidence interval of the change in its next-token KL divergence from the full model lay within 10\% of that
divergence, and its realized reads changed by less than 0.5 percentage points. For the full model, its KL
divergence from the single-GPU RTX PRO 6000 run stayed below 10\% of our reader's divergence. So the layout does
not materially change the numerics.

\section{Read accounting and profiler check}
\label{app:accounting}

For each token and MoE layer, our reader is charged for every weight and parameter byte it loads from memory.
This includes the ternary sketch codes and scales read during screening, the full-precision gate, up, and down
rows of the $k$ kept channels, the recovery map's coefficient vectors for the skipped channels and its shared
basis $U$, the precomputed per-channel norms, and the Hadamard rotation signs. Every read is counted in whole
32-byte hardware sectors at batch one, without modeling cache reuse across tokens. Every baseline is charged
under the same sector model, in the weight layout that needs the fewest reads, including its thresholds,
low-bit copies, and other metadata. For selectors with thresholds, we report the reads they realize on each
benchmark, counted on the GPU. Table~\ref{tab:ledger} breaks down our reader's reads by component at the four
main budgets.

\begin{table}[tb]
\centering
\caption{\textbf{Most of our reader's reads go to exact channels, and the sketch takes most of the rest.} Routed-expert reads per token at batch one, as a percentage of the full model, by component at the four main budgets.
At the two smallest budgets, the partial screen reads part of the ternary sketch.}
\label{tab:ledger}
\small
\begin{tabular}{lrlrrrr}
\toprule
 & Total reads & & \multicolumn{4}{c}{Of which (\%)} \\
\cmidrule(lr){4-7}
Model & (\%) & Sketch read & Sketch & Exact channels & Recovery & Metadata \\
\midrule
Qwen3-Thinking & 12.85 & partial & 4.86 & 7.29 & 0.66 & 0.04 \\
 & 17.83 & partial & 4.86 & 12.30 & 0.63 & 0.04 \\
 & 28.68 & full & 7.99 & 20.00 & 0.59 & 0.10 \\
 & 38.62 & full & 7.99 & 30.00 & 0.54 & 0.10 \\
\addlinespace
Gemma~4 & 12.84 & partial & 4.78 & 7.49 & 0.54 & 0.03 \\
 & 17.83 & partial & 4.78 & 12.50 & 0.52 & 0.03 \\
 & 28.42 & full & 7.86 & 19.99 & 0.49 & 0.08 \\
 & 38.40 & full & 7.86 & 30.01 & 0.45 & 0.07 \\
\addlinespace
Qwen3.5 & 12.84 & partial & 5.08 & 6.98 & 0.74 & 0.04 \\
 & 17.87 & partial & 5.08 & 12.04 & 0.72 & 0.04 \\
 & 29.11 & full & 8.33 & 20.00 & 0.68 & 0.11 \\
 & 39.06 & full & 8.33 & 30.00 & 0.62 & 0.10 \\
\bottomrule
\end{tabular}

\end{table}

\begin{table}[tb]
\centering
\caption{\textbf{Our reader reads part of the sketch at the two smallest budgets and all of it above.}
Each column is one budget. It gives the share of candidate channels computed exactly across each token's
routed experts and the resulting total reads, which include the ternary sketch, exact channels, recovery map,
and metadata, averaged over the three post-trained models and three benchmarks. Ranges give the lowest and
highest share across the three models.}
\label{tab:budgets}
\small
\begin{tabular*}{\linewidth}{@{\extracolsep{\fill}}lcccccc@{}}
\toprule
Reads (\% of full model) & 12.8 & 17.8 & 28.7 & 38.7 & 58.6 & 78.5 \\
\midrule
Sketch read & partial & partial & full & full & full & full \\
Channels computed exactly (\%) & 7.0--7.5 & 12.0--12.5 & 20 & 30 & 50 & 70 \\
\bottomrule
\end{tabular*}
\end{table}

To check that the sector count matches what the hardware actually transfers, we profiled 12 evenly spaced MoE layers of each post-trained model with NVIDIA Nsight Compute, out of 48 in Qwen3-Thinking, 30 in Gemma~4, and 40 in Qwen3.5.
The profiled kernels ran on weights, recovery maps, and hidden states captured from a live decoding step, with the cache flushed before each kernel launch.
Table~\ref{tab:ncu} compares the measured DRAM reads of the routed experts with our count.
Our reader's DRAM traffic is 18.3\% of the full model's at batch one on all three models, against the 17.8 to 17.9\% we charge.
The share rises at larger batches, to 25.7\% at batch eight on Qwen3-Thinking, because tokens in a batch may select different channels.
Counting once each sector that the batch's tokens select together predicts this rise: the measured share exceeds the count by at most 0.5 percentage points at every batch size.

The router, Gemma~4's dense MLP, and Qwen3.5's shared expert run in full in both versions.
Our reader's whole MoE block, with these modules included, reads 18.9\% of the full model's at batch one on Qwen3-Thinking, 41.0\% on Gemma~4, and 29.0\% on Qwen3.5.

\begin{table}[tb]
\centering
\caption{\textbf{The profiler measures 18.3\% of the full model's DRAM reads at batch one on all three models, within 0.5 percentage points of our count.} Our reader's DRAM reads as a percentage of the full model's, averaged over 12 MoE layers per model with a cold cache. The first two rows of each model cover the routed experts. Counted is the reads we charge at batch one and, at larger batches, the sectors the batch's tokens select together, each counted once. The third row adds the router, Gemma~4's dense MLP, and Qwen3.5's shared expert, which both versions run in full. Next to each model is the full model's routed-expert reads per layer at batch one.}
\label{tab:ncu}
\small
\begin{tabular}{lccc}
\toprule
Batch size & 1 & 4 & 8 \\
\midrule
\multicolumn{4}{l}{\emph{Qwen3-Thinking} (full model: 75.6\,MB per layer)} \\
Measured (\%) & 18.3 & 21.8 & 25.7 \\
Counted (\%) & 17.8 & 21.4 & 25.2 \\
Measured, whole MoE block (\%) & 18.9 & 22.0 & 25.8 \\
\midrule
\multicolumn{4}{l}{\emph{Gemma~4} (full model: 95.3\,MB per layer)} \\
Measured (\%) & 18.3 & 21.2 & 23.9 \\
Counted (\%) & 17.8 & 20.8 & 23.5 \\
Measured, whole MoE block (\%) & 41.0 & 30.2 & 29.4 \\
\midrule
\multicolumn{4}{l}{\emph{Qwen3.5} (full model: 50.4\,MB per layer)} \\
Measured (\%) & 18.3 & 19.4 & 20.4 \\
Counted (\%) & 17.9 & 18.9 & 20.0 \\
Measured, whole MoE block (\%) & 29.0 & 22.7 & 22.2 \\
\bottomrule
\end{tabular}
\end{table}

\section{Calibrated thresholds read less than their budget}
\label{app:drift}

Section~\ref{sec:drift} counts 72 benchmark runs: Prox and our reader's threshold version, each at four
budgets. Table~\ref{tab:drift} extends this count to every selector we ran with a calibrated threshold on the
post-trained models, 24 threshold settings on all nine model-benchmark pairs. All 216 runs read less than the
budget they were calibrated for. For Prox and our reader's threshold version, this budget is the read target
of the calibration fit. For TEAL and WINA's score with a threshold, it is the target share of inputs kept. For
CATS, it is the reads of its top-$k$ version at the same target sparsity.

\begin{table}[tb]
\centering
\caption{\textbf{Every calibrated threshold reads less than the budget it was calibrated for.} Each row is one selector at one calibrated budget.
The columns give the lowest, mean, and highest realized routed-expert reads (\%) over the nine model-benchmark pairs.
The budget is the read target for Prox and our reader's threshold version, the share of inputs kept for TEAL and
WINA's score, and the reads of the top-$k$ version for CATS; where it differs across pairs, the table shows its
mean, and each run is compared with its own budget.}
\label{tab:drift}
\small
\begin{tabular}{lrrrr}
\toprule
 & & \multicolumn{3}{c}{Realized reads (\%)} \\
\cmidrule(lr){3-5}
Method & Calibrated for (\%) & Min & Mean & Max \\
\midrule
CATS & 53.3 & 51.1 & 52.1 & 53.0 \\
 & 66.7 & 63.8 & 65.1 & 66.0 \\
\addlinespace
Our reader, calibrated & 12.8 & 11.2 & 11.8 & 12.4 \\
 & 17.8 & 15.2 & 16.5 & 17.1 \\
 & 28.7 & 24.5 & 26.5 & 27.7 \\
 & 38.7 & 32.9 & 35.8 & 37.2 \\
\addlinespace
Prox as published, calibrated & 13.0 & 5.6 & 8.5 & 10.1 \\
 & 18.0 & 9.9 & 12.9 & 14.7 \\
 & 28.0 & 18.9 & 22.5 & 24.6 \\
 & 38.0 & 29.6 & 32.6 & 34.1 \\
 & 58.0 & 47.3 & 49.8 & 51.0 \\
 & 78.0 & 67.1 & 70.0 & 71.4 \\
\addlinespace
TEAL as published, calibrated & 15.0 & 6.4 & 8.9 & 11.0 \\
 & 20.0 & 9.8 & 12.6 & 15.3 \\
 & 30.0 & 19.9 & 22.6 & 24.9 \\
 & 40.0 & 32.6 & 35.0 & 36.2 \\
 & 50.0 & 43.7 & 46.2 & 47.4 \\
 & 60.0 & 54.7 & 56.9 & 58.0 \\
\addlinespace
WINA score, calibrated & 15.0 & 5.4 & 7.8 & 9.6 \\
 & 20.0 & 8.2 & 11.2 & 13.9 \\
 & 30.0 & 17.2 & 20.7 & 23.5 \\
 & 40.0 & 32.0 & 34.3 & 35.8 \\
 & 50.0 & 43.5 & 46.1 & 47.7 \\
 & 60.0 & 54.8 & 57.1 & 58.2 \\
\bottomrule
\end{tabular}

\end{table}

\subsection{Calibrating on the model's own generations}
\label{app:own-generations}

UltraChat assistant turns contain no reasoning traces, so we also fitted Prox's thresholds on each full
model's own reasoning, for targets of 13.4, 15, 18.4, and 20\%. The full model wrote one thinking-mode trace,
up to 32{,}768 tokens, for each of 1{,}024 UltraChat first user turns. These turns lie outside every fitting,
tuning, and evaluation split, and none shares a 13-gram with a benchmark prompt. The traces average 1{,}669
generated tokens on Gemma~4 and 2{,}302 on Qwen3-Thinking. As in our chat calibration, Prox's recipe uses 30
of these sequences and 20{,}480 evenly spaced positions. Each sequence is a prompt plus its trace, cut at a
random point and capped at 8{,}192 tokens.

Prox with these trace-calibrated thresholds still reads less than its targets on Gemma~4 GPQA-Diamond (four
samples per question): 10.5, 11.6, 14.0, and 15.2\% of reads. It scores 0.0, 0.3, 23.9, and 45.1, and 100,
100, 65, and 35\% of its answers hit the 32{,}768-token limit. For comparison, Prox with its threshold
calibrated on chat for 18\% scores 11.7 at 13.3\% of reads. On Gemma~4 IFEval, Prox with a trace-calibrated
threshold scores 31.2 at 14.9\% of reads, against 11.3 with chat calibration.

On Qwen3-Thinking GPQA-Diamond, Prox with trace-calibrated thresholds scores 57.1 at 9.6\% of reads and 62.8 at
11.2\%, against 59.7 at 9.9\% with chat calibration. On IFEval it scores 82.1, against 82.8. Calibrating on
the model's own reasoning therefore neither stops the reads from drifting nor prevents Gemma~4's collapse into
repetition loops.

\section{What a calibrated threshold's channel count follows}
\label{app:allocation}

With a calibrated threshold, the number of channels kept changes from token to token. This appendix asks what
that number follows, and whether the variation spends reads where tokens need them. Unless stated otherwise,
we use Prox's INT4 score on Qwen3-Thinking with the thresholds from our main evaluation, calibrated on UltraChat
for 13\% and 18\% of reads. Where stated, we also use thresholds calibrated on the model's own reasoning
(Appendix~\ref{app:own-generations}). Gemma~4 repeats the context test, and Gemma~4 and Qwen3.5 repeat the
equal-read and generation tests, with thresholds calibrated on UltraChat. Every variant ranks channels by the
same score; only the rule that sets each token's count changes.

\paragraph{Setup.} We use 48 GPQA-Diamond reasoning traces written by the full Qwen3-Thinking model and
teacher-force 256-token windows starting at answer positions 0, 2{,}000, and 6{,}000. The full model processes the question and the earlier answer. Inside the window, each
variant runs on hidden states from its own sparse layers, and we measure its next-token KL divergence from the
full model. We compare the threshold with two controls. \emph{Per-layer fixed counts} give every token in a
layer the same count, set so that reads in the window match the threshold's; a two-pass procedure matches them
to a relative residual below 0.05\%. \emph{Shuffled counts} randomly permute the threshold's per-token counts
across positions. To measure how much each token needs extra channels, a separate single-step experiment scores
576 tokens, four positions in each of the three windows of 48 traces. Each token is scored after a full-model
prefix at 0.5, 0.7, 1, 1.4, and 2 times the top-$k$ count, and its \emph{need} is how much its KL divergence
falls from 0.5 to 2 times the count.

\paragraph{The earlier answer sets the count.} Table~\ref{tab:context-cut} takes the window at position 6{,}000
and changes only what comes before it. Removing the earlier answer, so that the window follows the question
directly, raises the kept count by $56.6\pm2.7\%$. This brings it to 89\% of the count at the start of the
answer, where the tokens themselves differ. Among identical token IDs, Prox keeps 41\% fewer channels at
position 6{,}000 than at the start, and the count falls equally in every band of the full model's next-token
loss. On Gemma~4, whose reads fall less (from 10.5\% to 9.7\%), removing the earlier answer raises the kept
count by $12.5\pm0.5\%$. With thresholds calibrated on Qwen3-Thinking's own traces, reads fall similarly, from
11.3\% to 7.7\% over 6{,}000 tokens. Both the length and the content of the earlier answer matter. With the
exact scores and chat-calibrated threshold of Figure~\ref{fig:drift}a, the window keeps $30.1\pm1.3\%$ more
channels after the question alone and $10.4\pm0.6\%$ more after the last 1{,}000 answer tokens. Placing an
answer to a different question, also 6{,}000 tokens long, before the window instead of the model's own answer
raises the count by $19.3\pm1.3\%$.

\begin{table}[tb]
\centering
\caption{\textbf{The same tokens keep more channels when the earlier answer is removed.} The first three rows
change only the context before the 256-token windows at answer position 6{,}000 of Qwen3-Thinking
GPQA-Diamond traces, under Prox's threshold calibrated for 13\%. Kept channels are per token, summed over 48
layers (of 294{,}912 candidates) and averaged over 48 traces; Change is the mean over traces of each trace's change from the first row; $\pm$ is one paired SE. The last row
shows the start of the answer, which has different tokens.}
\label{tab:context-cut}
\small
\begin{tabular}{lrr@{}l}
\toprule
Context before the window & Kept channels & \multicolumn{2}{c}{Change (\%)} \\
\midrule
Question and all 6{,}000 earlier answer tokens & 9{,}983 & \multicolumn{2}{c}{reference} \\
Question and the last 1{,}000 answer tokens & 11{,}423 & $+15.0$ & $\se{1.2}$ \\
Question only & 15{,}422 & $+56.6$ & $\se{2.7}$ \\
\midrule
Start of the answer (different tokens) & 17{,}283 & & \\
\bottomrule
\end{tabular}
\end{table}

\paragraph{What the count follows on Qwen3-Thinking.} In the single-step experiment, the threshold's per-token
count correlates with answer position at $-0.57$ (Spearman, at both budgets), with need at only $-0.13$
(13\%) and $-0.19$ (18\%), and with the full model's next-token loss at $-0.29$ and $-0.26$. Need itself
follows loss ($+0.58$, $+0.68$) rather than position ($+0.09$, $+0.14$). Even among tokens at the same
position, the count correlates slightly negatively with need ($-0.11\pm0.04$ and $-0.12\pm0.05$ over the 12
positions) and with loss ($-0.19\pm0.05$ and $-0.16\pm0.06$). On this model, removing the dependence on
answer position would therefore leave variation that slightly favors tokens with lower need. On each model's
own sparse states in the three windows, the count correlates with loss at $-0.16$ to $-0.23$ on
Qwen3-Thinking, $-0.21$ to $-0.28$ on Gemma~4, and $0.00$ to $+0.14$ on Qwen3.5. Yet on Qwen3.5, variable
counts still improve quality at equal reads (below), so loss alone does not decide whether the variation
helps.

\paragraph{Equal reads.} Does the variation help at equal reads? Pooled over the three windows, per-layer
fixed counts lower KL divergence relative to the chat-calibrated threshold by $3.3\pm1.0\%$ at 13\% and
$3.4\pm1.2\%$ at 18\%, and relative to the trace-calibrated threshold by $4.0\pm0.9\%$ and $2.9\pm1.2\%$. On
Gemma~4 at 13\%, they lower it by $12.5\pm1.5\%$. On Qwen3.5 they raise it, by $10.5\pm1.1\%$ at 13\% and
$14.8\pm1.2\%$ at 18\%, and the threshold is ahead in all three windows at both budgets. So on Qwen3-Thinking
and Gemma~4 reasoning the variation hurts, and on Qwen3.5 it helps. The threshold's counts still carry
information on every model: shuffling them across positions raises pooled KL on each. On held-out chat, the
distribution the threshold was calibrated on, fixed counts instead raise KL by $6.5\pm1.9\%$ and
$9.2\pm2.2\%$.

\paragraph{Moving reads across positions.} Making each layer's count constant across the three windows barely changes KL at equal pooled reads.
We compare per-window fixed counts, matched to the threshold version's reads in each window, with counts held at their three-window mean.
At equal pooled reads, the constant counts change KL by $-0.3\pm0.5\%$ and $-0.2\pm0.6\%$ on Qwen3-Thinking, $+0.1\pm0.8\%$ on
Gemma~4, and $+1.0\pm0.5\%$ (just beyond two SE) and $+0.1\pm0.7\%$ on Qwen3.5. A reversed schedule, which
reflects each layer's counts about their three-window mean, is worse:
$+6.8\pm0.7\%$ and $+7.1\pm1.1\%$ on Qwen3-Thinking, and $+4.2\pm0.7\%$ and $+3.5\pm0.8\%$ on Qwen3.5.

\paragraph{Predicting need.} An oracle that knows every token's loss could halve KL divergence at the same
total reads, but a predictor of need does not get close. A per-layer ridge regression from the token's position and
summary statistics of its screened channel scores (total and largest score mass, share in the top channels,
entropy, and count above the threshold) explains about 20\% of the variance in log KL. Our offline estimate of
the allocation it implies is worse than fixed counts.

\paragraph{Generation.} Qwen3-Thinking IFEval is the benchmark where a threshold beats top-$k$ at its
realized reads. There, Prox keeps its top-$k$ on inputs and uses a channel threshold calibrated for 13\%. With
this threshold, it scores 82.4 at 10.0\% of reads. With per-layer fixed counts matched to the threshold's
per-layer budgets, it scores 77.6 at 10.3\% ($-4.8\pm1.8$), and with the same count in every layer, 72.8 at
10.3\%. On GPQA-Diamond (one sample), it scores 59.1 at 9.9\% with the threshold and 56.6 at 9.9\% with fixed
counts ($-2.5\pm2.8$, a tie). Per-token counts thus help on IFEval but show no detectable gain on GPQA-Diamond.
IFEval answers are shorter than GPQA-Diamond reasoning and fail on a single violated instruction, which may
make a few tokens decisive; we did not test this.

On Qwen3.5, with its threshold calibrated for 13\%, Prox scores 89.5 at 9.7\% of reads on IFEval, against 88.5
at 9.7\% with fixed counts ($-0.9\pm1.3$, a tie). On GPQA-Diamond (one sample), it scores 79.8 at 9.4\%,
against 72.7 at 9.1\% ($-7.1\pm2.7$). Without per-token counts, Qwen3.5's answers grow longer, 16.2k against
11.3k tokens on average, and 5.1\% of them hit the length limit, against 1.0\%.

On Gemma~4, removing the per-token variation does not prevent the collapse. On IFEval, with per-layer fixed
counts matched to the threshold's per-layer budgets, Prox scores 0.0 at 10.1\% of reads, and 99.8\% of its
answers stop at the length limit. Prox's threshold version also scores 0.0 at 10.1\%; replacing only its
input threshold with top-$k$ gives 3.5 (Appendix~\ref{app:gemma}). With the same count in every layer, Prox's
top-$k$ version scores 16.5 at 11.7\%. On GPQA-Diamond, Prox with fixed counts also scores 0.0 at 9.8\% of
reads with every answer stopped at the limit, as does its threshold version (0.0 at 9.9\%). Gemma~4's collapse therefore comes with the threshold's calibrated
per-layer budgets or the low reads they settle at, rather than with per-token variation; this test cannot
separate the two.

\section{Every keep-rule comparison}
\label{app:keep-rule}

\begin{figure}[tb]
\centering
\includegraphics[width=\linewidth]{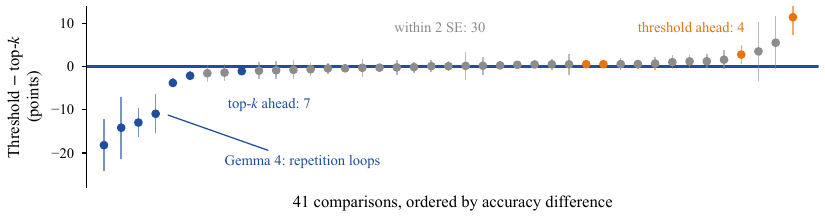}
\caption{\textbf{Figure~\ref{fig:ranked} with each top-$k$ version run at the reads its threshold version
actually used.}
Each dot is one comparison, as in Figure~\ref{fig:ranked}. When a threshold version read more than one
percentage point away from its calibrated budget, the top-$k$ version is rerun at those realized reads. The
figure has 41 comparisons at 13\% and 18\% budgets; four TEAL comparisons at heavy sparsity, where both
versions collapse, are left out. Threshold versions read less than their budget, so this rerun lowers the
top-$k$ version's reads too, yet the top-$k$ version matches or beats the threshold version in 37 of 41.}
\label{fig:ranked-realized}
\end{figure}

Tables~\ref{tab:keep-generation} and~\ref{tab:keep-likelihood} list every paired comparison of a selector's
threshold and top-$k$ versions, and Tables~\ref{tab:keep-generation-realized} and~\ref{tab:keep-likelihood-realized}
repeat them at realized reads; Figure~\ref{fig:ranked} summarizes the 43 at 13\% and 18\% budgets. Each row
fixes one channel score and compares the two versions at the budget the threshold was calibrated for. When the
threshold version's realized reads differ from that budget by more than one percentage point, the top-$k$
version is also rerun at those reads (Figure~\ref{fig:ranked-realized}).

In the 23 generation-task comparisons, the top-$k$ version leads beyond two standard errors in 8 and the
threshold version in none. Two TEAL comparisons at heavy sparsity, where both versions fall below 10\% of the
full model, are marked and not counted. We fixed this set of comparisons in advance, not by outcome: Prox's
score and its output-scaled variant on IFEval at 13\% and on GPQA-Diamond at 13\% for Gemma~4 and Qwen3-Thinking, TEAL's score on IFEval at 18\%, and
GSM8K and BBH on OLMoE at both budgets. Table~\ref{tab:headline} covers the remaining pairs for Prox: over all
nine model-benchmark pairs at 13\% and 18\%, its threshold version scores higher than its top-$k$ version in
2 of 18, both on Qwen3.5 GPQA-Diamond and by at most 1.5 points. In the 20 likelihood-task comparisons on OLMoE at 13\% and 18\%, the
top-$k$ version leads in 2 and the threshold version in 3. At the 38\% budget, Prox's top-$k$ version keeps
50\% of inputs and its threshold version 30\%, so the two differ in more than the keep rule. These 10
comparisons (top-$k$ version ahead in 3, threshold version in none) appear in parentheses and are left out of
every count. Table~\ref{tab:keep-pooled} pools each model's counted comparisons.

In the generation-task comparisons with Prox's score and its output-scaled variant, both versions select inputs
by top-$k$, so only the channel threshold is replaced. In the comparisons with TEAL's score, the threshold on
inputs is replaced by top-$k$.

\begin{table}[tb]
\centering
\caption{\textbf{At the calibrated budget, top-$k$ leads in 8 of 23 generation comparisons and the threshold
in none.} Each row fixes one model, task, score, and budget, and compares the accuracy of the threshold and
top-$k$ versions at the budget the threshold was calibrated for. The difference is threshold minus top-$k$
in accuracy points, $\pm$ one paired standard error. \emph{Ahead} names the version ahead by more than two standard errors; $=$ marks a
tie, and -- marks a comparison where both versions fall below 10\% of the full model, left out of the count.}
\label{tab:keep-generation}
\small
\setlength{\tabcolsep}{4pt}
\begin{tabular}{llllrrr@{}lc}
\toprule
 &  &  &  & \multicolumn{2}{c}{Accuracy (\%)} & & & \\
\cmidrule(lr){5-6}
Model & Task & Score & Budget (\%) & Threshold & Top-$k$ & \multicolumn{2}{c}{Difference} & Ahead \\
\midrule
Gemma~4 & GPQA-D & Prox, output-scaled & 13 & 16.2 & 54.0 & $-37.9$ & $\se{3.8}$ & top-$k$ \\
 &  & Prox & 13 & 6.1 & 51.0 & $-44.9$ & $\se{3.5}$ & top-$k$ \\
 & IFEval & Prox, output-scaled & 13 & 23.1 & 63.6 & $-40.5$ & $\se{2.3}$ & top-$k$ \\
 &  & Prox & 13 & 3.5 & 45.8 & $-42.3$ & $\se{2.2}$ & top-$k$ \\
 &  & TEAL & 18 & 0.0 & 7.0 & $-7.0$ & $\se{1.1}$ & -- \\
\addlinespace
OLMoE & BBH & Prox, output-scaled & 13 & 25.4 & 26.1 & $-0.7$ & $\se{0.6}$ & = \\
 &  & Prox & 13 & 26.1 & 26.2 & $-0.1$ & $\se{0.6}$ & = \\
 &  & TEAL & 13 & 3.0 & 5.4 & $-2.4$ & $\se{0.3}$ & top-$k$ \\
 &  & Prox, output-scaled & 18 & 28.5 & 30.1 & $-1.6$ & $\se{0.6}$ & top-$k$ \\
 &  & Prox & 18 & 28.1 & 28.5 & $-0.4$ & $\se{0.5}$ & = \\
 &  & TEAL & 18 & 10.8 & 17.6 & $-6.8$ & $\se{0.5}$ & top-$k$ \\
 & GSM8K & Prox, output-scaled & 13 & 9.6 & 10.4 & $-0.8$ & $\se{1.0}$ & = \\
 &  & Prox & 13 & 8.3 & 9.9 & $-1.5$ & $\se{0.9}$ & = \\
 &  & TEAL & 13 & 1.8 & 2.2 & $-0.4$ & $\se{0.4}$ & = \\
 &  & Prox, output-scaled & 18 & 12.1 & 11.0 & $+1.1$ & $\se{1.0}$ & = \\
 &  & Prox & 18 & 9.6 & 11.0 & $-1.4$ & $\se{0.9}$ & = \\
 &  & TEAL & 18 & 2.1 & 2.1 & $+0.0$ & $\se{0.5}$ & = \\
\addlinespace
Qwen3.5 & IFEval & Prox, output-scaled & 13 & 91.1 & 91.9 & $-0.7$ & $\se{1.0}$ & = \\
 &  & Prox & 13 & 91.5 & 90.9 & $+0.6$ & $\se{1.1}$ & = \\
 &  & TEAL & 18 & 3.0 & 10.4 & $-7.4$ & $\se{1.5}$ & top-$k$ \\
\addlinespace
Qwen3-Thinking & GPQA-D & Prox, output-scaled & 13 & 62.6 & 59.6 & $+3.0$ & $\se{3.0}$ & = \\
 &  & Prox & 13 & 57.1 & 58.1 & $-1.0$ & $\se{3.0}$ & = \\
 & IFEval & Prox, output-scaled & 13 & 85.4 & 85.6 & $-0.2$ & $\se{1.4}$ & = \\
 &  & Prox & 13 & 84.3 & 85.4 & $-1.1$ & $\se{1.6}$ & = \\
 &  & TEAL & 18 & 0.0 & 0.0 & $+0.0$ & $\se{0.0}$ & -- \\
\bottomrule
\end{tabular}

\end{table}

\begin{table}[tb]
\centering
\caption{\textbf{At realized reads, top-$k$ leads in 5 of 21 generation comparisons and the threshold in 1.}
Columns follow Table~\ref{tab:keep-generation}, with each top-$k$ version run at its threshold version's
realized reads. The top-$k$ version is rerun only where the
threshold version's realized reads differ from the budget by more than one percentage point. $^\dagger$The threshold
version read within one point of its budget, so the top-$k$ run at the budget is reused.}
\label{tab:keep-generation-realized}
\small
\setlength{\tabcolsep}{3.5pt}
\begin{tabular}{llllrrr@{}lc}
\toprule
 &  &  &  & \multicolumn{2}{c}{Accuracy (\%)} & & & \\
\cmidrule(lr){5-6}
Model & Task & Score & Budget (\%) & Threshold & Top-$k$ & \multicolumn{2}{c}{Difference} & Ahead \\
\midrule
Gemma~4 & GPQA-D & Prox, output-scaled & 13 & 16.2 & 30.3 & $-14.1$ & $\se{3.5}$ & top-$k$ \\
 &  & Prox & 13 & 6.1 & 24.2 & $-18.2$ & $\se{2.9}$ & top-$k$ \\
 & IFEval & Prox, output-scaled & 13 & 23.1 & 34.0 & $-10.9$ & $\se{2.1}$ & top-$k$ \\
 &  & Prox & 13 & 3.5 & 16.5 & $-12.9$ & $\se{1.6}$ & top-$k$ \\
 &  & TEAL & 18 & 0.0 & 0.2 & $-0.2$ & $\se{0.2}$ & -- \\
\addlinespace
OLMoE & BBH & Prox, output-scaled & 13 & 25.4 & 25.3 & $+0.1$ & $\se{0.6}$ & = \\
 &  & Prox & 13 & 26.1 & 26.2\rlap{$^\dagger$} & $-0.1$ & $\se{0.6}$ & = \\
 &  & TEAL & 13 & 3.0 & 3.0 & $+0.1$ & $\se{0.2}$ & -- \\
 &  & Prox, output-scaled & 18 & 28.5 & 28.0 & $+0.6$ & $\se{0.6}$ & = \\
 &  & Prox & 18 & 28.1 & 28.5\rlap{$^\dagger$} & $-0.4$ & $\se{0.5}$ & = \\
 &  & TEAL & 18 & 10.8 & 12.9 & $-2.1$ & $\se{0.4}$ & top-$k$ \\
 & GSM8K & Prox, output-scaled & 13 & 9.6 & 10.4\rlap{$^\dagger$} & $-0.8$ & $\se{1.0}$ & = \\
 &  & Prox & 13 & 8.3 & 9.9\rlap{$^\dagger$} & $-1.5$ & $\se{0.9}$ & = \\
 &  & TEAL & 13 & 1.8 & 2.2\rlap{$^\dagger$} & $-0.4$ & $\se{0.4}$ & = \\
 &  & Prox, output-scaled & 18 & 12.1 & 11.8 & $+0.2$ & $\se{1.0}$ & = \\
 &  & Prox & 18 & 9.6 & 11.0\rlap{$^\dagger$} & $-1.4$ & $\se{0.9}$ & = \\
 &  & TEAL & 18 & 2.1 & 1.5 & $+0.6$ & $\se{0.4}$ & = \\
\addlinespace
Qwen3.5 & IFEval & Prox, output-scaled & 13 & 91.1 & 91.9\rlap{$^\dagger$} & $-0.7$ & $\se{1.0}$ & = \\
 &  & Prox & 13 & 91.5 & 90.9\rlap{$^\dagger$} & $+0.6$ & $\se{1.1}$ & = \\
 &  & TEAL & 18 & 3.0 & 0.0 & $+3.0$ & $\se{0.7}$ & -- \\
\addlinespace
Qwen3-Thinking & GPQA-D & Prox, output-scaled & 13 & 62.6 & 57.1 & $+5.6$ & $\se{3.0}$ & = \\
 &  & Prox & 13 & 57.1 & 53.5 & $+3.5$ & $\se{3.3}$ & = \\
 & IFEval & Prox, output-scaled & 13 & 85.4 & 85.2 & $+0.2$ & $\se{1.5}$ & = \\
 &  & Prox & 13 & 84.3 & 72.8 & $+11.5$ & $\se{1.9}$ & threshold \\
 &  & TEAL & 18 & 0.0 & 0.0\rlap{$^\dagger$} & $+0.0$ & $\se{0.0}$ & -- \\
\bottomrule
\end{tabular}

\end{table}

\begin{table}[tb]
\centering
\caption{\textbf{On Prox's five likelihood tasks on OLMoE, the two versions tie in 15 of 20 comparisons;
top-$k$ leads in 2 and the threshold in 3.} Prox's threshold and top-$k$ versions, with the prompt either sparse
or computed by the full model. Columns follow
Table~\ref{tab:keep-generation}. At 38\%, the top-$k$ version keeps 50\% of inputs and the threshold version
30\%; these verdicts are in parentheses and not counted.}
\label{tab:keep-likelihood}
\small
\setlength{\tabcolsep}{4pt}
\begin{tabular}{lllrrr@{}lc}
\toprule
 &  &  & \multicolumn{2}{c}{Accuracy (\%)} & & & \\
\cmidrule(lr){4-5}
Prompt & Task & Budget (\%) & Threshold & Top-$k$ & \multicolumn{2}{c}{Difference} & Ahead \\
\midrule
full model & ARC-C & 13 & 49.7 & 50.0 & $-0.3$ & $\se{0.9}$ & = \\
 & ARC-E & 13 & 73.9 & 73.2 & $+0.7$ & $\se{0.6}$ & = \\
 & HellaSwag & 13 & 75.4 & 74.9 & $+0.6$ & $\se{0.3}$ & threshold \\
 & MMLU & 13 & 50.3 & 50.3 & $-0.0$ & $\se{0.3}$ & = \\
 & PIQA & 13 & 77.7 & 76.4 & $+1.3$ & $\se{0.7}$ & = \\
\addlinespace
 & ARC-C & 18 & 51.9 & 51.4 & $+0.5$ & $\se{0.9}$ & = \\
 & ARC-E & 18 & 75.8 & 75.8 & $+0.0$ & $\se{0.6}$ & = \\
 & HellaSwag & 18 & 77.2 & 76.6 & $+0.6$ & $\se{0.2}$ & threshold \\
 & MMLU & 18 & 50.4 & 50.6 & $-0.2$ & $\se{0.3}$ & = \\
 & PIQA & 18 & 78.6 & 77.5 & $+1.1$ & $\se{0.6}$ & = \\
\addlinespace
 & ARC-C & 38 & 55.1 & 55.3 & $-0.2$ & $\se{0.7}$ & (=) \\
 & ARC-E & 38 & 77.8 & 78.6 & $-0.8$ & $\se{0.4}$ & (=) \\
 & HellaSwag & 38 & 78.7 & 78.9 & $-0.2$ & $\se{0.2}$ & (=) \\
 & MMLU & 38 & 51.6 & 52.1 & $-0.5$ & $\se{0.3}$ & (=) \\
 & PIQA & 38 & 79.2 & 80.1 & $-0.9$ & $\se{0.5}$ & (=) \\
\addlinespace
sparse & ARC-C & 13 & 44.6 & 43.0 & $+1.6$ & $\se{1.0}$ & = \\
 & ARC-E & 13 & 71.4 & 70.3 & $+1.1$ & $\se{0.7}$ & = \\
 & HellaSwag & 13 & 73.6 & 73.3 & $+0.3$ & $\se{0.3}$ & = \\
 & MMLU & 13 & 35.7 & 39.5 & $-3.8$ & $\se{0.4}$ & top-$k$ \\
 & PIQA & 13 & 76.0 & 76.1 & $-0.2$ & $\se{0.8}$ & = \\
\addlinespace
 & ARC-C & 18 & 49.6 & 46.8 & $+2.8$ & $\se{1.0}$ & threshold \\
 & ARC-E & 18 & 74.9 & 73.7 & $+1.2$ & $\se{0.6}$ & = \\
 & HellaSwag & 18 & 76.4 & 76.0 & $+0.4$ & $\se{0.3}$ & = \\
 & MMLU & 18 & 42.1 & 43.1 & $-1.0$ & $\se{0.4}$ & top-$k$ \\
 & PIQA & 18 & 77.8 & 78.4 & $-0.6$ & $\se{0.7}$ & = \\
\addlinespace
 & ARC-C & 38 & 55.2 & 54.4 & $+0.8$ & $\se{0.8}$ & (=) \\
 & ARC-E & 38 & 77.4 & 77.9 & $-0.5$ & $\se{0.5}$ & (=) \\
 & HellaSwag & 38 & 78.8 & 79.4 & $-0.6$ & $\se{0.2}$ & (top-$k$) \\
 & MMLU & 38 & 49.5 & 50.7 & $-1.1$ & $\se{0.3}$ & (top-$k$) \\
 & PIQA & 38 & 79.0 & 80.1 & $-1.1$ & $\se{0.5}$ & (top-$k$) \\
\bottomrule
\end{tabular}

\end{table}

\begin{table}[tb]
\centering
\caption{\textbf{At realized reads, top-$k$ again leads in 2 of 20 likelihood comparisons and the threshold
in 3.} Columns follow Table~\ref{tab:keep-generation}, with each top-$k$ version run at its threshold version's
realized reads. $^\dagger$The threshold version read within one
point of its budget, so the top-$k$ run at the budget is reused.}
\label{tab:keep-likelihood-realized}
\small
\setlength{\tabcolsep}{4pt}
\begin{tabular}{lllrrr@{}lc}
\toprule
 &  &  & \multicolumn{2}{c}{Accuracy (\%)} & & & \\
\cmidrule(lr){4-5}
Prompt & Task & Budget (\%) & Threshold & Top-$k$ & \multicolumn{2}{c}{Difference} & Ahead \\
\midrule
full model & ARC-C & 13 & 49.7 & 50.0\rlap{$^\dagger$} & $-0.3$ & $\se{0.9}$ & = \\
 & ARC-E & 13 & 73.9 & 73.2\rlap{$^\dagger$} & $+0.7$ & $\se{0.6}$ & = \\
 & HellaSwag & 13 & 75.4 & 74.9\rlap{$^\dagger$} & $+0.6$ & $\se{0.3}$ & threshold \\
 & MMLU & 13 & 50.3 & 49.7 & $+0.5$ & $\se{0.3}$ & = \\
 & PIQA & 13 & 77.7 & 76.4\rlap{$^\dagger$} & $+1.3$ & $\se{0.7}$ & = \\
\addlinespace
 & ARC-C & 18 & 51.9 & 52.8 & $-0.9$ & $\se{0.9}$ & = \\
 & ARC-E & 18 & 75.8 & 75.7 & $+0.1$ & $\se{0.5}$ & = \\
 & HellaSwag & 18 & 77.2 & 76.6\rlap{$^\dagger$} & $+0.6$ & $\se{0.2}$ & threshold \\
 & MMLU & 18 & 50.4 & 50.6 & $-0.2$ & $\se{0.3}$ & = \\
 & PIQA & 18 & 78.6 & 78.0 & $+0.6$ & $\se{0.6}$ & = \\
\addlinespace
 & ARC-C & 38 & 55.1 & 54.9 & $+0.2$ & $\se{0.8}$ & (=) \\
 & ARC-E & 38 & 77.8 & 78.7 & $-1.0$ & $\se{0.4}$ & (top-$k$) \\
 & HellaSwag & 38 & 78.7 & 79.0 & $-0.2$ & $\se{0.2}$ & (=) \\
 & MMLU & 38 & 51.6 & 51.8 & $-0.2$ & $\se{0.3}$ & (=) \\
 & PIQA & 38 & 79.2 & 79.7 & $-0.4$ & $\se{0.5}$ & (=) \\
\addlinespace
sparse & ARC-C & 13 & 44.6 & 43.0\rlap{$^\dagger$} & $+1.6$ & $\se{1.0}$ & = \\
 & ARC-E & 13 & 71.4 & 70.3\rlap{$^\dagger$} & $+1.1$ & $\se{0.7}$ & = \\
 & HellaSwag & 13 & 73.6 & 73.3\rlap{$^\dagger$} & $+0.3$ & $\se{0.3}$ & = \\
 & MMLU & 13 & 35.7 & 39.5\rlap{$^\dagger$} & $-3.8$ & $\se{0.4}$ & top-$k$ \\
 & PIQA & 13 & 76.0 & 76.1\rlap{$^\dagger$} & $-0.2$ & $\se{0.8}$ & = \\
\addlinespace
 & ARC-C & 18 & 49.6 & 46.8\rlap{$^\dagger$} & $+2.8$ & $\se{1.0}$ & threshold \\
 & ARC-E & 18 & 74.9 & 73.7\rlap{$^\dagger$} & $+1.2$ & $\se{0.6}$ & = \\
 & HellaSwag & 18 & 76.4 & 76.0\rlap{$^\dagger$} & $+0.4$ & $\se{0.3}$ & = \\
 & MMLU & 18 & 42.1 & 43.1\rlap{$^\dagger$} & $-1.0$ & $\se{0.4}$ & top-$k$ \\
 & PIQA & 18 & 77.8 & 78.4\rlap{$^\dagger$} & $-0.6$ & $\se{0.7}$ & = \\
\addlinespace
 & ARC-C & 38 & 55.2 & 54.9 & $+0.3$ & $\se{0.8}$ & (=) \\
 & ARC-E & 38 & 77.4 & 77.9\rlap{$^\dagger$} & $-0.5$ & $\se{0.5}$ & (=) \\
 & HellaSwag & 38 & 78.8 & 79.1 & $-0.3$ & $\se{0.2}$ & (=) \\
 & MMLU & 38 & 49.5 & 50.7\rlap{$^\dagger$} & $-1.1$ & $\se{0.3}$ & (top-$k$) \\
 & PIQA & 38 & 79.0 & 80.3 & $-1.3$ & $\se{0.5}$ & (top-$k$) \\
\bottomrule
\end{tabular}

\end{table}

\begin{table}[tb]
\centering
\caption{\textbf{No model favors the threshold version at the calibrated budget.} Threshold-version
minus top-$k$-version accuracy, pooled over each model's counted comparisons with inverse-variance weights, at
the calibrated budget and at realized reads. The $\pm$ is one SE that treats comparisons as independent;
accounting for shared questions changes each SE by at most 0.1. The realized-read columns favor the threshold
version (Section~\ref{sec:prelim}). Only Qwen3-Thinking leads beyond two SE there, a lead that comes from Prox
on IFEval and reverses at the calibrated budget.}
\label{tab:keep-pooled}
\small
\begin{tabular}{lrr@{}lrr@{}l}
\toprule
 & \multicolumn{3}{c}{At the calibrated budget} & \multicolumn{3}{c}{At realized reads} \\
\cmidrule(lr){2-4}\cmidrule(lr){5-7}
Model & Comparisons & \multicolumn{2}{c}{Threshold $-$ top-$k$} & Comparisons & \multicolumn{2}{c}{Threshold $-$ top-$k$} \\
\midrule
Qwen3-Thinking & 4 & $-0.3$ & $\se{1.0}$ & 4 & $+4.5$ & $\se{1.1}$ \\
Qwen3.5 & 3 & $-1.7$ & $\se{0.7}$ & 2 & $-0.2$ & $\se{0.8}$ \\
OLMoE & 32 & $-0.3$ & $\se{0.1}$ & 31 & $+0.0$ & $\se{0.1}$ \\
Gemma~4 & 4 & $-41.5$ & $\se{1.3}$ & 4 & $-13.3$ & $\se{1.1}$ \\
\bottomrule
\end{tabular}

\end{table}

\subsection{Applying Prox to OLMoE}
\label{app:prox-olmoe}

Prox \citep{liu2026prox} was evaluated on dense models, and five of its six tasks are multiple-choice
likelihood tasks. We apply Prox to the routed experts of OLMoE-1B-7B-0924 and evaluate the same five tasks in lm-eval
0.4.3 \citep{gao2024lmeval} (Table~\ref{tab:prox-tasks}). Each task runs under two prompt settings. In the first, as in Prox, the
whole sequence is sparse. In the second, the full model computes the prompt except its last token, so only the
positions that predict an answer token are sparse.

We fit the thresholds with Prox's sequential procedure. Layers are calibrated in order, with all earlier layers
already sparse, and each layer gets one input threshold and one channel threshold from empirical quantiles of
up to 200{,}000 activation samples pooled across its routed experts. Three settings differ from Prox's dense
setup. First, OLMoE is a base model, so we calibrate on 20{,}480 token positions from 30 Pile documents instead
of Alpaca text. Second, we set the budgets (13, 18, and 38\%) in charged sector reads instead of through Prox's
latency model. Third, the threshold version's input threshold is fitted to keep 30\% of inputs at all three
budgets, while the top-$k$ version keeps 30\% of inputs at 13\% and 18\% and 50\% at 38\%.

At realized reads and the 13\% and 18\% budgets, with the whole sequence sparse as in Prox, the threshold
version leads in 1 of 10 comparisons and the top-$k$ version in 2. With the prompt computed by the full model,
the threshold version leads in 2 and the top-$k$ version in none (Tables~\ref{tab:keep-likelihood} and~\ref{tab:keep-likelihood-realized}).

\begin{table}[tb]
\centering
\caption{\textbf{Prox's five likelihood tasks on OLMoE.} Few-shot examples, metric, number of questions, and full-model accuracy for each task.}
\label{tab:prox-tasks}
\small
\begin{tabular}{lcccr}
\toprule
Task & Shots & Metric & Questions & Full-model accuracy (\%) \\
\midrule
MMLU & 5 & accuracy & 14{,}042 & 52.71 \\
HellaSwag & 10 & normalized accuracy & 10{,}042 & 79.40 \\
ARC-E & 0 & accuracy & 2{,}376 & 78.11 \\
ARC-C & 25 & normalized accuracy & 1{,}172 & 55.20 \\
PIQA & 0 & accuracy & 1{,}838 & 79.98 \\
\bottomrule
\end{tabular}

\end{table}

\subsection{Repetition loops on Gemma~4}
\label{app:gemma}

Prox selects in two stages: a threshold on input coordinates, then a threshold on expert channels.
Section~\ref{sec:collapse} showed that Gemma~4 falls into repetition loops under Prox's thresholds even at
nearly equal reads. To find which stage causes the collapse, we replaced one threshold at a time with top-$k$,
at the 13\% budget, on the first GPQA-Diamond sample of each question (realized reads in parentheses). With
both thresholds, Prox scores 0.0 (10.9\% of reads), and every answer hits the length limit. With top-$k$ on
inputs and the threshold on channels, it scores 6.1 (11.4\%). With the threshold on inputs and top-$k$ on
channels, it scores 43.9 (12.4\%), and with top-$k$ at both stages, 51.0 (13.0\%). On IFEval, the same four
settings score 0.0, 3.5, 43.1, and 45.8.

Replacing the channel threshold with top-$k$ therefore raises GPQA-Diamond accuracy by 43.9 to 44.9 points
while adding only 1.5 to 1.6 percentage points of reads, and raises IFEval accuracy by 42.3 to 43.1 points.
Replacing the input threshold changes accuracy by at most 7.1 points on either benchmark. So replacing the
channel threshold, not the input threshold, is what avoids the collapse. Giving every token in a layer the same
count, set to the average the threshold kept in that layer, does not avoid it either
(Appendix~\ref{app:allocation}); that test cannot tell whether the calibrated layer budgets or the low reads
they settle at drive the collapse.

The collapse is not limited to 13\%: with its threshold calibrated for 18\%, Prox finishes its reasoning in
only 132 of 792 GPQA-Diamond answers, against 787 of 792 for the full model. At the 13\% budget, a per-token floor on
channels does not prevent the collapse either. When every token keeps at least 4\% of its candidate channels on top of the calibrated
threshold, Prox still scores only 2.0 on GPQA-Diamond (11.1\% of reads). Finally, when set to keep all
channels, our implementations of Prox and TEAL match the full model on held-out chat, which rules out a gross
implementation error in these baselines.

\subsection{Our reader with a calibrated threshold, and CATS}
\label{app:calibrated-reader}

\paragraph{Our reader.} To test the keep-rule swap on our own reader, we replace its per-token top-$k$ with one
threshold per layer on the same score. The threshold is fitted on UltraChat with Prox's recipe, so that it
keeps $k$ channels per token on average. On the benchmarks, this threshold version reads 0.4 to 5.8 percentage
points less than its budget. Of 36 paired comparisons, across three post-trained models, three benchmarks, and
four budgets, 34 are ties and none favors the threshold version. Both remaining comparisons favor the top-$k$
version, on Gemma~4 at 12.8\% of reads. There the threshold version reads 11.7\% and scores 62.4 against 69.4
on GPQA-Diamond ($-7.1\pm1.7$) and 73.2 against 80.4 on MMLU-Pro ($-7.2\pm1.2$). It also hits the length limit
on 15 to 17\% of answers, against 4 to 6\% for the top-$k$ version.

\paragraph{CATS.} CATS computes the gate projection exactly and keeps channels by gate magnitude. We compare its
threshold and top-$k$ versions at two budgets, about 53\% and 67\% of reads, on all nine model-benchmark
pairs. Of the 18 comparisons, 16 are ties. The threshold version leads in the other two, both on Gemma~4 at
53\%: by $5.8\pm1.7$ points on GPQA-Diamond and $5.5\pm1.1$ points on MMLU-Pro, while reading slightly less
(52.7\% and 52.8\%, against 53.3\%).

\subsection{Calibrated layer budgets and calibrated recovery}
\label{app:calibrated-parts}

\paragraph{Layer budgets.} We compare our reader's uniform layer budget with non-uniform per-layer budgets,
derived either from the weights or from calibration data, at the same total reads of 17.8\%. Calibrated minus
uniform, accuracy changes by $-0.76\pm1.10$ on GPQA-Diamond (four samples) and $+1.11\pm1.01$ on IFEval on
Gemma~4, and by $-1.77\pm1.18$ and $-1.85\pm1.31$ on Qwen3-Thinking. On OLMoE, calibrated layer budgets lower
next-token KL divergence by about 0.02 on held-out Pile text from the calibration distribution. Yet they change
GSM8K accuracy by at most 0.1 points and BBH accuracy by $-1.23\pm0.50$ and $-0.28\pm0.50$ at the two smallest
budgets. The KL gain is not confined to the calibration data. On Qwen3-Thinking with our reader at 17.8\% of
reads, we measure teacher-forced KL divergence on 144 windows from the full model's GPQA-Diamond answers,
with every layer sparse inside each window, fitting budgets on half of the answers and evaluating on the other
half. Relative to the uniform budget, the chat-calibrated budgets lower KL
divergence by $15.7\pm1.9\%$, close to the $18.7\pm2.1\%$ of budgets fitted on GPQA-Diamond answers, while
budgets derived from the weights change it by $+2.4\pm1.7\%$. Lower KL divergence on reasoning text thus did
not turn into a detected benchmark gain.

\paragraph{Recovery.} We also fit the low-rank recovery map on calibration text by minimizing next-token KL
divergence from the full model, at the same reads as our weight-only map. We make 18 comparisons on the three post-trained models:
GPQA-Diamond, IFEval, MMLU-Pro, and GSM8K at 17.8\% of reads, plus IFEval and MMLU-Pro at 12.8\%. The
calibrated map never beats the weight-only map beyond two standard errors, and the weight-only map leads on
Qwen3-Thinking MMLU-Pro at 12.8\% ($+2.1\pm1.0$). On OLMoE, the calibrated map lowers next-token KL divergence
on held-out Pile text from its calibration distribution (0.158 against 0.200 at 17.8\% of reads) but gives no
benchmark gain beyond two standard errors.

\FloatBarrier
\section{What each part of the reader contributes}
\label{app:components}

\paragraph{Score with the full sketch.} When the reader reads the full ternary sketch (from 28.7\% of reads
upward), each 128-coordinate block also comes with the mean-squared quantization error $\rho_b$ of its ternary
code. This gives each screened inner product an error variance $\sigma^2=\sum_b\rho_b\|(Rx)_b\|^2$ (all norms are Euclidean). We model
the true gate and up pre-activations as independent Gaussians centered at $\hat g_j$ and $\hat v_j$, with
variance $\sigma^2$ computed from each row's own residuals. Five-node Gauss--Hermite quadrature then gives the
mean $\mu_j$ and second moment $m_j$ of the gated activation $a_j=\phi(g_j^\top x)(v_j^\top x)$. Let
$n_j=\|d_j\|$, and let $q_j=Ub_j$ be channel $j$'s low-rank recovery vector, with component
$\beta_j^\parallel=d_j^\top q_j/n_j$ along $d_j$ and orthogonal remainder of norm $\beta_j^\perp=\|q_j-\beta_j^\parallel d_j/n_j\|$. We score channel
$j$ by the expected squared output error of skipping it after recovery,
\begin{equation*}
  s_j=w_j^2\left[n_j^2(m_j-\mu_j^2)+(n_j\mu_j-\beta_j^\parallel p_j)^2+(\beta_j^\perp p_j)^2\right].
\end{equation*}

\paragraph{Output energy and KL divergence.} Next-token KL divergence closely follows how much of the routed
output's energy the kept channels preserve, $E=1-\|y-y_{\mathcal{K}}\|^2/\|y\|^2$. Here $y$ is the full
routed-expert output and $y_{\mathcal{K}}$ the output of the kept channels alone. We measure $E$ on all 16
OLMoE layers, at 3{,}000 positions per layer, for four scores under per-token top-$k$, each at four read
budgets without recovery. The four scores are our output-scaled ternary score, two versions of Prox's score
using each token's largest 30\% and 50\% of input coordinates, and an INT4 screen of the gate projection. Across these 16 settings,
$E$ and held-out next-token KL divergence have a rank correlation of $-0.96$, and KL divergence scales roughly
as $(1-E)^{1.2\text{--}1.9}$.

\paragraph{Ablations.} Table~\ref{tab:components} compares variants of our reader on 192 held-out OLMoE
documents, by next-token KL divergence from the full model at equal total reads.

\begin{table}[tb]
\centering
\caption{\textbf{Partial screening helps at small budgets, and recovery helps at every budget.}
Next-token KL divergence from the full model on held-out OLMoE documents at equal reads (lower is better). Each
row is one variant of our reader; $^\star$ marks the variant our reader uses at each budget.}
\label{tab:components}
\small
\begin{tabular}{lcccc}
\toprule
Reads (\%) & 13.5 & 18.5 & 28.5 & 38.4 \\
\midrule
Full sketch, no recovery & 0.553 & 0.277 & 0.107 & 0.051 \\
Full sketch with recovery & 0.488 & 0.245 & 0.096$^\star$ & 0.047$^\star$ \\
Partial sketch with recovery & 0.310$^\star$ & 0.187$^\star$ & 0.090 & 0.053 \\
Prox's 4-bit copy, output-scaled top-$k$, recovery & 0.305 & 0.187 & 0.070 & 0.035 \\
\bottomrule
\end{tabular}
\end{table}

Our reader uses the partial sketch at the two smallest budgets and the full sketch at higher budgets. We also
replace our ternary sketch (1.0\,GB next to the OLMoE checkpoint) with Prox's 4-bit weight copy (2.16\,GB),
keeping our output-scaled top-$k$ and weight-only recovery. This variant also needs no calibration data. It
matches our reader at the two smallest budgets and has lower KL divergence from 28.5\% of reads upward, at
about twice the resident auxiliary memory.

\paragraph{Quantized checkpoints.} Our reader also works on quantized checkpoints. On OLMoE held-out documents,
with reads charged in BF16-equivalent routed bytes including quantization scales, our reader on a GPTQ \citep{frantar2023gptq} 4-bit
checkpoint reaches a next-token KL divergence of 0.045 at 18.9\% of reads. AWQ \citep{lin2024awq} 3-bit weights alone reach 0.075
at 19.7\% (paired difference $-0.030\pm0.003$). At 21.5\% of reads our reader reaches 0.033, against 0.019 for
GPTQ 4-bit weights alone at 26.0\%. On a quantized checkpoint, the ternary sketch takes a larger share of
resident memory: a 4-bit checkpoint plus the sketch occupies about 34\% of the BF16 routed bytes, against 20\%
for 3-bit weights.

\paragraph{Design of the screen.} Table~\ref{tab:design} tests the screen's design choices on development
data. Removing the Hadamard rotation, or forcing each routed expert to keep a fixed share of channels, raises
KL divergence on OLMoE. For the partial screen, we chose to read 80 of every 128 rotated coordinates per block
on OLMoE development data. There, 80 gives the lowest KL divergence, or lies within about 0.004 of it, at each
of three budgets. On Gemma~4 development data, 80 is also lowest at two budgets and within 0.0003 of the lowest
at the third.

\begin{table}[tb]
\centering
\caption{\textbf{Rotation and ranking channels across all routed experts lower KL divergence, and reading 80 of
128 coordinates is at or near the best.} Next-token KL divergence from the full model on development
data (lower is better). Top: at 17.8\% of reads on OLMoE without recovery, the full sketch, two of its design variants, and two other
scores under top-$k$. Bottom: coordinates
the partial screen reads per 128-coordinate block, with recovery, at equal reads; bold marks the lowest in each
row.}
\label{tab:design}
\small
\begin{tabular}{lc}
\toprule
Variant at 17.8\% of reads (OLMoE, development) & KL \\
\midrule
Full sketch, no recovery (rotated, top-$k$ over all routed experts) & 0.286 \\
Without rotation & 0.348 \\
Fixed share of channels per expert & 0.491 \\
\addlinespace
Score from an INT4 copy of the gate projection, same top-$k$ & 0.811 \\
Score from TEAL's input magnitudes, top-$k$ inputs & 1.471 \\
\bottomrule
\end{tabular}

\vspace{6pt}

\begin{tabular}{llcccccc}
\toprule
& & \multicolumn{6}{c}{Coordinates read per 128-coordinate block} \\
\cmidrule(lr){3-8}
Model & Budget (\% of reads) & 128 & 96 & 80 & 64 & 48 & 32 \\
\midrule
OLMoE & 13 & 0.443 & 0.336 & \textbf{0.316} & 0.317 & 0.343 & 0.424 \\
 & 18 & 0.224 & 0.198 & \textbf{0.194} & 0.198 & 0.229 & 0.297 \\
 & 28 & 0.102 & \textbf{0.094} & 0.098 & 0.106 & 0.127 & 0.173 \\
\addlinespace
Gemma~4 & 13 & 0.235 & 0.188 & \textbf{0.176} & 0.177 & 0.197 & 0.260 \\
 & 18 & 0.124 & 0.110 & \textbf{0.106} & 0.111 & 0.129 & 0.179 \\
 & 28 & 0.056 & \textbf{0.052} & \textbf{0.052} & 0.057 & 0.070 & 0.103 \\
\bottomrule
\end{tabular}

\end{table}

\paragraph{Sensitivity to rank, score, and sketch precision.} Table~\ref{tab:sensitivity} varies three of our reader's settings on OLMoE.
Panels (a) and (b) compare each pair of variants at equal reads on the 192 held-out documents of Appendix~\ref{app:olmoe-fidelity}.
A recovery map of rank 32 is best or tied at each of the three budgets tested.
Lower ranks save too few reads to make up for the skipped output they leave unrecovered.
Higher ranks spend reads that do more good as extra exact channels.
The differences are small: except for rank 128 at the two lowest budgets, every rank stays within 3.6\% of rank 32's KL divergence.
The choice of score matters more.
With the full sketch, the point score of Equation~\ref{eq:score} has 16.5\% and 10.0\% lower KL divergence than the expected-error score above at 13.5\% and 18.5\% of reads.
The two scores are level at 28.5\%, and the point score has 14.2\% higher KL divergence at 38.4\%.
Our reader uses each score where it is better or level: the point score at its two lowest budgets and the expected-error score from 28.7\% upward.

Panel (c) measures sketch precision by output energy, as defined above: each rotated sketch ranks channels by the point score and picks 10\% of them, and these channels are then computed exactly.
More precise sketches carry more of the routed output's energy, while they take more bytes.
We did not test the 2-bit sketch in decoding.
Prox's 4-bit weight copy, used with our top-$k$ and recovery, tests a 4-bit screen: it matches our reader at the two smallest budgets and has lower KL divergence from 28.5\% of reads upward, at about twice the resident memory (Table~\ref{tab:components}).

\begin{table}[tb]
\centering
\caption{\textbf{At equal reads, rank 32 is best or tied at every budget tested, and our reader uses each score where it is better or level.} Panels (a) and (b) give the relative change in next-token KL divergence from a reference, $100\,(\mathrm{KL}-\mathrm{KL}_{\mathrm{ref}})/\mathrm{KL}_{\mathrm{ref}}$, at equal reads on 192 held-out OLMoE documents, with one paired standard error; positive means higher KL divergence than the reference, which is worse. Ranks 64 and 128 read 0.6 to 1.8 percentage points more than rank 32, so each is compared with rank 32 given the same reads. Panel (c) gives the output energy $E$ carried by the 10\% of channels each screen selects, computed exactly, on 16 OLMoE layers at 3{,}000 development positions each, and the screen's size as a share of the routed experts' BF16 bytes. Selecting with exact activations instead of a screen carries 86.4\%.}
\label{tab:sensitivity}
\small
\emph{(a) Recovery rank: change in KL divergence versus rank 32 (\%)}\\[2pt]
\begin{tabular}{lccc}
\toprule
Reads of rank 32 (\%) & 12.8 & 17.8 & 28.5 \\
\midrule
Rank 8 & $+$2.5 $\pm$ 0.9 & $+$2.4 $\pm$ 1.0 & $+$3.1 $\pm$ 1.0 \\
Rank 16 & $+$2.6 $\pm$ 1.0 & $-$0.7 $\pm$ 1.4 & $+$3.4 $\pm$ 0.7 \\
Rank 64 & $+$3.6 $\pm$ 1.0 & $+$1.7 $\pm$ 1.1 & $-$0.2 $\pm$ 1.2 \\
Rank 128 & $+$11.1 $\pm$ 1.2 & $+$6.7 $\pm$ 1.2 & $+$1.7 $\pm$ 0.9 \\
\bottomrule
\end{tabular}

\vspace{0.9em}
\emph{(b) Score with the full sketch: point score versus expected-error score, change in KL divergence (\%)}\\[2pt]
\begin{tabular}{lcccc}
\toprule
Reads (\%) & 13.5 & 18.5 & 28.5 & 38.4 \\
\midrule
Point score & $-$16.5 $\pm$ 0.9 & $-$10.0 $\pm$ 1.2 & $+$0.3 $\pm$ 1.4 & $+$14.2 $\pm$ 2.5 \\
\bottomrule
\end{tabular}

\vspace{0.9em}
\emph{(c) Sketch precision}\\[2pt]
\begin{tabular}{lcccc}
\toprule
Screen & Binary & Ternary (ours) & 2-bit & 4-bit \\
\midrule
Output energy $E$ carried (\%) & 75.2 & 81.0 & 83.2 & 86.1 \\
Size (\% of routed bytes) & 4.7 & 7.3 & 8.9 & 17.2 \\
\bottomrule
\end{tabular}
\end{table}

\FloatBarrier
\section{Every model and task}
\label{app:headline-grid}

\begin{figure}[tb]
\centering
\includegraphics[width=\linewidth]{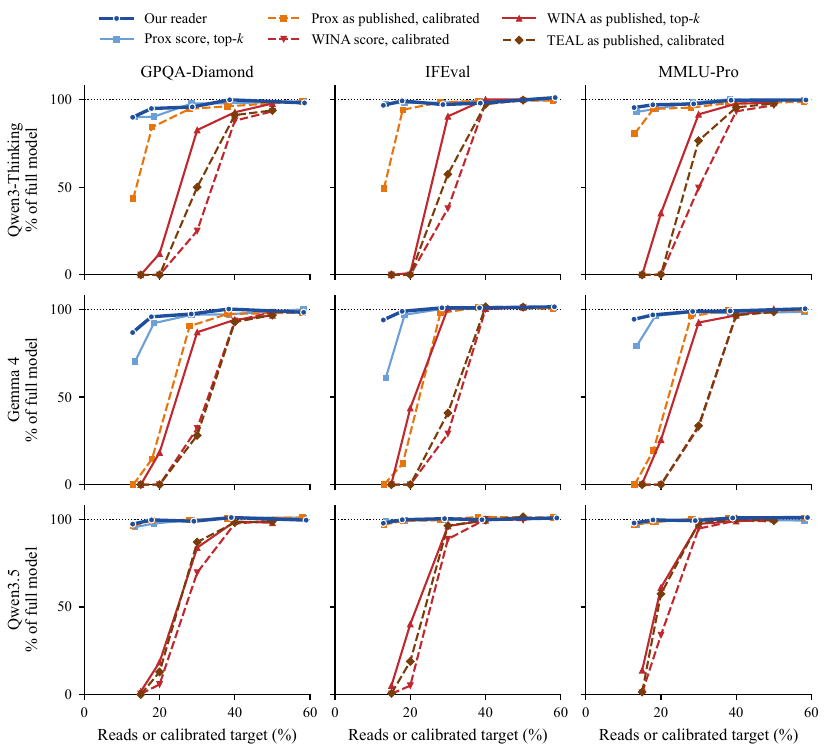}
\caption{\textbf{Accuracy against reads on every model and task.}
Each panel is one model-benchmark pair, and its vertical axis runs from zero to above the full model's accuracy
(dotted line). Blue solid lines use no calibration data. Warm dashed lines are selectors with thresholds fitted
on calibration data, placed at the budget they were calibrated for. WINA as published (red solid) keeps a top-$k$
per token, with a sparsity level for each projection fitted on calibration data as in TEAL. GPQA-Diamond
averages four samples per question, IFEval reports prompt-level strict accuracy, and MMLU-Pro uses a fixed
subset of 1{,}000 questions.}
\label{fig:headline-grid}
\end{figure}

\begin{figure}[tb]
\centering
\includegraphics[width=\linewidth]{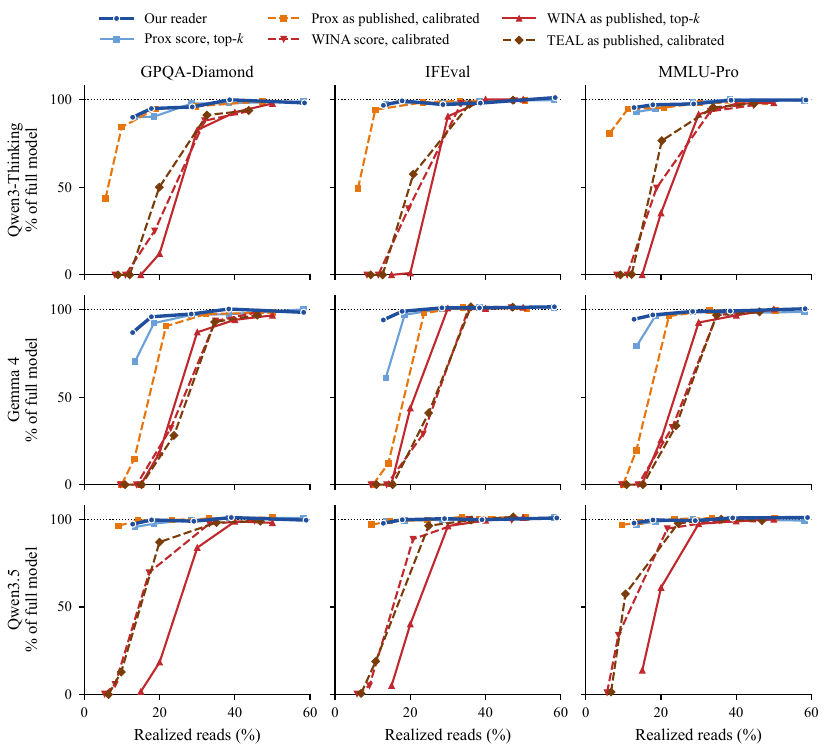}
\caption{\textbf{Figure~\ref{fig:headline-grid} with calibrated selectors placed at the reads they actually
used.}
Selectors with calibrated thresholds move left, because they read less at test time than the budget they were
calibrated for.}
\label{fig:headline-grid-realized}
\end{figure}

Figures~\ref{fig:headline-grid} and~\ref{fig:headline-grid-realized} show accuracy against reads for every
model and benchmark, with calibrated selectors at their calibrated budgets and at their realized reads.
Table~\ref{tab:headline} lists the scores at heavy sparsity. In a rerun of GPQA-Diamond with eight samples per
question, our reader at 17.8\% of reads trails the full model by $3.9\pm1.0$ points on Gemma~4 and
$4.6\pm1.4$ points on Qwen3-Thinking.

\paragraph{Answer length.} Our reader does not trade speed per token for longer answers. At 17.8\% of reads,
its mean answer length stays within about 15\% of the full model's on every model-benchmark pair. On
Qwen3-Thinking, Gemma~4, and Qwen3.5, it changes by $-3$, $+3$, and $+1\%$ on GPQA-Diamond, $+15$, $+9$, and
$-12\%$ on IFEval, and $-14$, $+14$, and $-10\%$ on MMLU-Pro. At most 1.3\% of its answers stop at the length
limit on any pair, and the same holds for the full model.

\begin{figure}[tb]
\centering
\includegraphics[width=\linewidth]{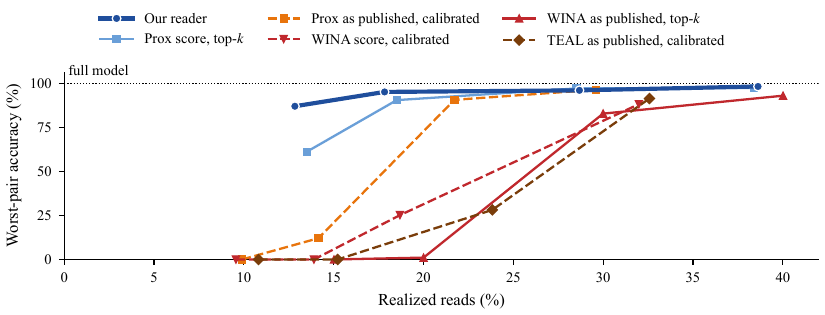}
\caption{\textbf{Figure~\ref{fig:headline} with calibrated selectors placed at the reads they actually used.}
Each curve is the worst accuracy over the nine model-benchmark pairs, as in Figure~\ref{fig:headline}, but each
calibrated selector sits at the mean reads it used on the pair that gives its worst accuracy, instead of the
budget it was calibrated for. Selectors with calibrated thresholds read less than that budget (Section~\ref{sec:drift}), so
their points move left.}
\label{fig:worst-cell-realized}
\end{figure}
\begin{table}[tb]
\centering
\caption{\textbf{Accuracy of every selector on every model and task.}
Accuracy in percent (higher is better) at heavy sparsity. Budget is the mean reads for selectors without calibration data and the
calibrated budget for the others. GPQA-Diamond (GPQA-D) averages four samples per question.}
\label{tab:headline}
\small
\begin{tabular}{lrrrr}
\toprule
Method & Budget (\%) & GPQA-D & IFEval & MMLU-Pro \\
\midrule
\multicolumn{5}{l}{\textit{Qwen3-Thinking}} \\
\quad Full model & 100 & 70.8 & 87.8 & 80.5 \\
\addlinespace[1.5pt]
\quad Our reader & 12.8 & 63.8 & 85.0 & 76.9 \\
 & 17.8 & 67.3 & 87.2 & 78.2 \\
 & 28.7 & 67.9 & 85.4 & 78.6 \\
 & 38.7 & 70.8 & 86.1 & 80.3 \\
\addlinespace[1.5pt]
\quad Prox score, top-$k$ & 13.5 & 63.8 & 85.8 & 74.9 \\
 & 18.5 & 64.0 & 86.3 & 76.3 \\
 & 28.5 & 69.3 & 86.3 & 79.3 \\
\addlinespace[1.5pt]
\quad Prox as published, calibrated & 13 & 30.8 & 43.3 & 64.9 \\
 & 18 & 59.7 & 82.8 & 76.3 \\
 & 28 & 67.2 & 86.5 & 76.9 \\
\addlinespace[1.5pt]
\quad WINA as published, top-$k$ & 20 & 8.6 & 0.9 & 28.4 \\
 & 30 & 58.6 & 79.5 & 73.8 \\
\addlinespace[1.5pt]
\quad TEAL as published, calibrated & 20 & 0.0 & 0.0 & 0.2 \\
 & 30 & 35.5 & 50.5 & 61.7 \\
\addlinespace[1.5pt]
\quad WINA score, calibrated & 20 & 0.0 & 0.0 & 0.0 \\
 & 30 & 17.8 & 33.3 & 40.0 \\
\midrule
\multicolumn{5}{l}{\textit{Gemma~4}} \\
\quad Full model & 100 & 79.9 & 93.5 & 85.1 \\
\addlinespace[1.5pt]
\quad Our reader & 12.8 & 69.4 & 88.0 & 80.4 \\
 & 17.8 & 76.6 & 92.6 & 82.5 \\
 & 28.7 & 77.9 & 94.5 & 84.2 \\
 & 38.7 & 80.2 & 94.5 & 84.4 \\
\addlinespace[1.5pt]
\quad Prox score, top-$k$ & 13.5 & 56.3 & 57.1 & 67.3 \\
 & 18.5 & 73.7 & 90.8 & 82.0 \\
 & 28.5 & 77.5 & 93.9 & 83.9 \\
\addlinespace[1.5pt]
\quad Prox as published, calibrated & 13 & 0.0 & 0.0 & 0.0 \\
 & 18 & 11.7 & 11.3 & 16.7 \\
 & 28 & 72.3 & 91.7 & 81.9 \\
\addlinespace[1.5pt]
\quad WINA as published, top-$k$ & 20 & 14.6 & 40.9 & 21.8 \\
 & 30 & 69.6 & 94.3 & 78.8 \\
\addlinespace[1.5pt]
\quad TEAL as published, calibrated & 20 & 0.0 & 0.0 & 0.0 \\
 & 30 & 22.5 & 38.3 & 28.6 \\
\addlinespace[1.5pt]
\quad WINA score, calibrated & 20 & 0.0 & 0.0 & 0.0 \\
 & 30 & 25.8 & 27.2 & 27.9 \\
\midrule
\multicolumn{5}{l}{\textit{Qwen3.5}} \\
\quad Full model & 100 & 83.6 & 92.1 & 85.5 \\
\addlinespace[1.5pt]
\quad Our reader & 12.8 & 81.3 & 90.0 & 83.7 \\
 & 17.8 & 83.3 & 91.9 & 85.2 \\
 & 28.7 & 82.7 & 92.4 & 84.9 \\
 & 38.7 & 84.5 & 91.9 & 86.2 \\
\addlinespace[1.5pt]
\quad Prox score, top-$k$ & 13.5 & 79.9 & 91.3 & 83.0 \\
 & 18.5 & 81.6 & 91.5 & 84.4 \\
 & 28.5 & 83.3 & 91.7 & 85.5 \\
\addlinespace[1.5pt]
\quad Prox as published, calibrated & 13 & 80.6 & 89.3 & 82.8 \\
 & 18 & 83.1 & 91.1 & 84.1 \\
 & 28 & 83.3 & 91.7 & 85.7 \\
\addlinespace[1.5pt]
\quad WINA as published, top-$k$ & 20 & 15.4 & 37.0 & 52.2 \\
 & 30 & 70.1 & 88.5 & 83.3 \\
\addlinespace[1.5pt]
\quad TEAL as published, calibrated & 20 & 10.7 & 17.4 & 49.1 \\
 & 30 & 72.7 & 88.7 & 83.6 \\
\addlinespace[1.5pt]
\quad WINA score, calibrated & 20 & 4.8 & 4.6 & 29.0 \\
 & 30 & 58.2 & 81.7 & 81.0 \\
\bottomrule
\end{tabular}
\end{table}

\FloatBarrier
\section{Fidelity comparison with baselines}
\label{app:olmoe-baselines}

\subsection{OLMoE}
\label{app:olmoe-fidelity}

Table~\ref{tab:olmoe-baselines} compares our reader with nine prior channel selectors on OLMoE, by next-token
KL divergence from the full model and top-1 token agreement on 192 held-out Pile documents (12{,}096 decoded
positions), with the prompt processed by the full model. Calibrated baselines are fitted on disjoint Pile text
for budgets near our reader's and reported at the reads they realize. Prox divides each read budget between
its input screen and the channels it computes; we report both Prox's published division and the division with
the lowest KL divergence on development data. At each budget, our reader has lower KL divergence than every
baseline evaluated there, and it reads less than Prox's nearest point at the 13\% and 18\% budgets. Among
baselines evaluated at 13\% to 18\% of reads, every one except Prox has at least four times our reader's KL
divergence.

\begin{table}[!htb]
\centering
\caption{\textbf{On OLMoE, our reader is closer to the full model than every baseline evaluated at each budget.}
Next-token KL divergence from the full model (lower is better) and top-1 agreement on 192 held-out Pile
documents, at the reads each method realizes. HiRE-style screens channels with an INT4 copy of the gate
projection under our top-$k$. R-Sparse adds a low-rank term for skipped inputs. CHESS, CATS, M-COUNTDOWN (the
up-projection variant of COUNTDOWN \citep{cheon2025countdown}), and \citet{park2026intraexpert} compute one projection in full
before thresholding, so they read at least a third of each routed expert.}
\label{tab:olmoe-baselines}
\small
\begin{tabular}{llrcc}
\toprule
Budget & Method & Reads (\%) & KL ($\downarrow$) & \begin{tabular}[b]{@{}c@{}}Top-1\\agreement ($\uparrow$)\end{tabular} \\
\midrule
${\sim}13\%$ & Our reader & 12.8 & 0.334 & 0.751 \\
 & Prox, calibrated, division tuned on development data & 13.2 & 0.488 & 0.709 \\
 & Prox as published, calibrated & 13.7 & 0.462 & 0.723 \\
 & HiRE-style gate screen & 12.8 & 1.709 & 0.469 \\
\midrule
${\sim}18\%$ & Our reader & 17.8 & 0.199 & 0.808 \\
 & Prox, calibrated, division tuned on development data & 18.3 & 0.298 & 0.774 \\ & TEAL as published, calibrated & 18.5 & 1.109 & 0.613 \\
 & WINA score, calibrated & 17.6 & 1.295 & 0.583 \\
 & R-Sparse & 18.1 & 1.132 & 0.577 \\
 & HiRE-style gate screen & 17.8 & 0.860 & 0.602 \\
\midrule
${\sim}28\%$ & Our reader & 28.5 & 0.096 & 0.859 \\
 & Prox as published, calibrated & 29.1 & 0.136 & 0.842 \\ & TEAL as published, calibrated & 29.1 & 0.378 & 0.760 \\
 & WINA score, calibrated & 28.5 & 0.461 & 0.731 \\
 & R-Sparse & 28.8 & 0.445 & 0.717 \\
 & HiRE-style gate screen & 27.8 & 0.376 & 0.720 \\
\midrule
${\sim}38\%$ & Our reader & 38.4 & 0.047 & 0.900 \\
 & Prox, calibrated, division tuned on development data & 38.7 & 0.057 & 0.891 \\
 & Prox as published, calibrated & 38.3 & 0.060 & 0.892 \\
 & HiRE-style gate screen & 37.8 & 0.203 & 0.783 \\
 & CHESS & 40.4 & 0.643 & 0.663 \\
 & CATS & 40.5 & 1.113 & 0.555 \\
 & M-COUNTDOWN & 40.9 & 1.123 & 0.612 \\
 & \citet{park2026intraexpert} & 40.4 & 1.120 & 0.557 \\
\bottomrule
\end{tabular}
\end{table}

\subsection{Fresh held-out panels on the post-trained models}
\label{app:fresh-panels}

The post-trained models' chat panels were used during method development, so we also score each post-trained model on two new panels, drawn after every setting of our reader and of the baselines was fixed.
The first panel holds 192 UltraChat chat documents.
The second panel holds 64 thinking-mode traces that the full model wrote for UltraChat prompts, separate from the traces used for calibration in Appendix~\ref{app:own-generations}.
We score each trace over three 512-token windows, starting at its 0th, 500th, and 1{,}000th token.
We measure next-token KL divergence from the full model as on OLMoE, with the prompt processed by the full model.

Prox's calibrated thresholds realize different reads on each panel.
To compare at equal reads, we interpolate Prox's KL divergence for each document at our reader's reads, linearly in log KL divergence between the two Prox settings whose reads bracket ours.
Standard errors come from 2{,}000 paired resamples of documents.

Our reader has lower KL divergence than Prox at 17.8\% of reads on all 12 model-panel pairs, beyond two standard errors (Table~\ref{tab:fresh-panels}).
Our reader is lower than Prox at 28.5\% of reads on 8 pairs and level on 4.
Our reader is lower than Prox at 12.8\% of reads on all four Qwen3.5 panels.
No Prox setting reads this little on Gemma~4 or Qwen3-Thinking.
Our reader is still lower on every Gemma~4 panel against Prox's lowest setting, which reads 2.3 to 4.9 percentage points more.
Prox's lowest setting is lower than our reader on three of four Qwen3-Thinking panels.
At 38.5\%, our reader reads more than Prox's highest setting, so the panels allow no comparison at equal reads there.

The same panels repeat two component results on new data.
The partial screen has lower KL divergence than the full screen at both of its budgets on every panel, while reading less.
The recovery map fitted on calibration data (Appendix~\ref{app:calibrated-parts}) lowers KL divergence at 38.5\% of reads by 0.0003 to 0.010, beyond two standard errors on 11 of 12 panels.

\begin{table}[!htb]
\centering
\caption{\textbf{Our reader is closer to the full model than Prox at 17.8\% of reads on every model and fresh panel.} Our reader's next-token KL divergence minus Prox's, in units of $10^{-3}$, with one paired standard error; negative favors our reader. The chat panel holds 192 held-out UltraChat documents; each trace panel holds 512-token windows of the full model's own thinking traces, starting at the given token. Prox's value is interpolated at our reader's reads, which vary by at most 0.2 percentage points across models. $^\dagger$No Prox setting reads this little; the cell compares with Prox's lowest setting, which reads 2.3 to 4.9 points more.}
\label{tab:fresh-panels}
\small
\begin{tabular}{llrrr}
\toprule
& & \multicolumn{3}{c}{Our reader's reads (\%)} \\
\cmidrule(lr){3-5}
Model & Panel & 12.8 & 17.8 & 28.5 \\
\midrule
Qwen3-Thinking & Chat & $+$23.5 $\pm$ 4.2$^\dagger$ & $-$30.6 $\pm$ 2.7 & $-$5.4 $\pm$ 1.3 \\
 & Trace, from token 0 & $+$22.6 $\pm$ 9.2$^\dagger$ & $-$123.6 $\pm$ 7.0 & $-$20.0 $\pm$ 3.0 \\
 & Trace, from token 500 & $+$26.6 $\pm$ 14.0$^\dagger$ & $-$98.4 $\pm$ 10.7 & $-$17.7 $\pm$ 5.1 \\
 & Trace, from token 1{,}000 & $+$67.4 $\pm$ 10.7$^\dagger$ & $-$62.5 $\pm$ 8.8 & $-$7.4 $\pm$ 3.9 \\
\midrule
Gemma~4 & Chat & $-$20.7 $\pm$ 4.2$^\dagger$ & $-$95.7 $\pm$ 4.7 & $-$38.3 $\pm$ 2.2 \\
 & Trace, from token 0 & $-$180.6 $\pm$ 12.2$^\dagger$ & $-$180.2 $\pm$ 10.2 & $-$60.9 $\pm$ 3.4 \\
 & Trace, from token 500 & $-$148.4 $\pm$ 10.7$^\dagger$ & $-$172.7 $\pm$ 10.1 & $-$59.4 $\pm$ 3.4 \\
 & Trace, from token 1{,}000 & $-$92.0 $\pm$ 11.1$^\dagger$ & $-$205.1 $\pm$ 14.0 & $-$81.6 $\pm$ 5.6 \\
\midrule
Qwen3.5 & Chat & $-$12.9 $\pm$ 1.1 & $-$8.1 $\pm$ 0.8 & $-$1.4 $\pm$ 0.5 \\
 & Trace, from token 0 & $-$22.5 $\pm$ 1.9 & $-$12.0 $\pm$ 1.1 & 0.0 $\pm$ 0.6 \\
 & Trace, from token 500 & $-$19.8 $\pm$ 1.5 & $-$13.2 $\pm$ 1.1 & $-$0.8 $\pm$ 0.7 \\
 & Trace, from token 1{,}000 & $-$17.4 $\pm$ 2.0 & $-$11.0 $\pm$ 1.3 & $-$0.8 $\pm$ 0.8 \\
\bottomrule
\end{tabular}
\end{table}

\FloatBarrier
\section{Speed details}
\label{app:speed}

\paragraph{Setup.} We measure decoding speed for Qwen3-Thinking, Gemma~4, and Qwen3.5 at 17.8\% of
routed-expert reads, on a single RTX PRO 6000 Blackwell Max-Q GPU (300\,W) with vLLM 0.29.0, PyTorch 2.13, and
Triton 3.7.1. The full model runs vLLM's fused-MoE Triton kernels. At each batch size, we take the faster of
vLLM's default kernel configuration and one tuned for this GPU and model. Our reader and the full model use the same unmodified router.

\paragraph{Always-active modules.} Gemma~4's dense MLP and Qwen3.5's shared expert are part of the timed block.
Our implementation runs them on a second CUDA stream, alongside the router and the reader. Unmodified vLLM
does not overlap Gemma~4's dense MLP this way. If it did, Gemma~4's MoE-block speedup would be 1.61 to
2.19$\times$ instead of 1.71 to 2.34$\times$.

\paragraph{Strength of the baseline.} Unmodified vLLM is a strong baseline. At batch one, its fused-MoE expert
kernels stream weights at 1.35 to 1.44\,TB/s, which is 83\% to 88\% of the 1.63\,TB/s read bandwidth we measured
as achievable on this GPU.

\paragraph{MoE block.} To time the MoE blocks alone, we capture the layer inputs from seven real decoding steps
on 128-token chat prompts and replay all MoE layers, routing included, in one CUDA graph. After four warmup
replays, each process times seven repetitions with CUDA events and keeps the median. We report the median over
three fresh processes.

\paragraph{Whole model.} For whole-model timing, vLLM generates 64 output tokens from 128-token prompts at batch
sizes 1 to 128, with full CUDA graphs, one warmup run, and three timed repetitions in each of three
processes. Time per output token is the wall-clock time between the first and last generated tokens divided by
63, and we report the median across processes. This time includes attention, the language-model head,
sampling, and vLLM's scheduler, but not prompt prefill.

\begin{table}[tb]
\centering
\caption{\textbf{Speedups peak at batch sizes four to sixteen, and the whole-model gain fades by batch 128.} Decode times at 17.8\% of routed-expert reads, in milliseconds per decoding step, median of three processes.
Speedup is the full model's time divided by our reader's.
The MoE block covers all MoE layers with the router, Gemma~4's dense MLP, and Qwen3.5's shared expert.
Whole-model times use 128-token prompts.}
\label{tab:speed}
\footnotesize
\setlength{\tabcolsep}{3.5pt}
\begin{tabular}{llrrrrrrrr}
\toprule
\multicolumn{2}{l}{Batch size} & 1 & 2 & 4 & 8 & 16 & 32 & 64 & 128 \\
\midrule
\multicolumn{10}{l}{\textit{Qwen3-Thinking}} \\
\quad MoE block & Full model (ms) & 3.084 & 5.525 & 8.517 & 11.969 & 18.370 & \multicolumn{3}{c}{} \\
 & Our reader (ms) & 1.619 & 2.604 & 3.562 & 5.131 & 9.260 & \multicolumn{3}{c}{\textit{not measured}} \\
 & Speedup ($\times$) & 1.90 & 2.12 & 2.39 & 2.33 & 1.98 & \multicolumn{3}{c}{} \\
\quad Per output token & Full model (ms) & 6.253 & 9.454 & 12.450 & 15.666 & 22.044 & 27.366 & 34.107 & 38.196 \\
 & Our reader (ms) & 4.605 & 6.609 & 7.453 & 8.957 & 12.850 & 19.831 & 28.925 & 38.363 \\
 & Speedup ($\times$) & 1.36 & 1.43 & 1.67 & 1.75 & 1.72 & 1.38 & 1.18 & 1.00 \\
\midrule
\multicolumn{10}{l}{\textit{Gemma 4}} \\
\quad MoE block & Full model (ms) & 3.646 & 5.399 & 8.010 & 10.750 & 15.244 & \multicolumn{3}{c}{} \\
 & Our reader (ms) & 2.136 & 2.704 & 3.428 & 4.846 & 7.591 & \multicolumn{3}{c}{\textit{not measured}} \\
 & Speedup ($\times$) & 1.71 & 2.00 & 2.34 & 2.22 & 2.01 & \multicolumn{3}{c}{} \\
\quad Per output token & Full model (ms) & 6.459 & 8.260 & 11.060 & 14.005 & 19.122 & 23.250 & 29.890 & 36.345 \\
 & Our reader (ms) & 4.915 & 5.691 & 6.515 & 8.058 & 11.109 & 16.385 & 24.408 & 33.937 \\
 & Speedup ($\times$) & 1.31 & 1.45 & 1.70 & 1.74 & 1.72 & 1.42 & 1.22 & 1.07 \\
\midrule
\multicolumn{10}{l}{\textit{Qwen3.5}} \\
\quad MoE block & Full model (ms) & 2.163 & 3.511 & 5.503 & 9.132 & 15.085 & \multicolumn{3}{c}{} \\
 & Our reader (ms) & 1.350 & 2.049 & 2.193 & 3.394 & 5.760 & \multicolumn{3}{c}{\textit{not measured}} \\
 & Speedup ($\times$) & 1.60 & 1.71 & 2.51 & 2.69 & 2.62 & \multicolumn{3}{c}{} \\
\quad Per output token & Full model (ms) & 5.862 & 8.141 & 10.174 & 14.127 & 20.397 & 29.029 & 40.668 & 53.605 \\
 & Our reader (ms) & 5.092 & 6.680 & 6.943 & 8.540 & 10.837 & 16.680 & 30.812 & 43.353 \\
 & Speedup ($\times$) & 1.15 & 1.22 & 1.47 & 1.65 & 1.88 & 1.74 & 1.32 & 1.24 \\
\bottomrule
\end{tabular}

\end{table}

\paragraph{Same outputs.} The timed Triton kernels compute the same function as the evaluated reader, up to the
order of floating-point operations. On held-out chat documents at equal reads on Qwen3-Thinking, the two
implementations' next-token KL divergences from the full model differ by $-0.0011\pm0.0013$ (90\% confidence
interval). On all three models, the kernels keep the same channels as the evaluated reader at equal reads: 0 of
25{,}606 checked selection decisions differ.

\paragraph{Long contexts.} In single-process runs at 8{,}192 to 32{,}768 context tokens and batch sizes one and
four, whole-model decoding is 1.20 to 1.54$\times$ faster on Qwen3-Thinking, 1.16 to 1.48$\times$ on Gemma~4,
and 1.08 to 1.40$\times$ on Qwen3.5. We leave out batch four at 32{,}768 tokens on Gemma~4 and Qwen3.5, where
neither version fits in memory. The gain shrinks at longer contexts, because attention takes longer while the
MoE work per token stays the same.

\paragraph{Where the time goes.} Traces of whole decoding steps at batch one show that each Qwen3-Thinking MoE
layer takes 64.7\,$\mu$s in the full model and 34.4\,$\mu$s with our reader, router included. Within our
reader, screening takes 9.5\,$\mu$s, top-$k$ selection 5.0\,$\mu$s, and computing the kept gate, up, and down
rows 16.4\,$\mu$s. Projecting the skipped channels onto the rank-32 recovery map takes 8.0\,$\mu$s, partly
overlapped on a second CUDA stream. Attention takes 48\,$\mu$s per layer in both, so the whole model speeds up
less than the MoE block.

\begin{table}[tb]
\centering
\caption{\textbf{Our reader is faster than Park et al.'s kernels in the MoE block up to batch 32, and in the whole model at every batch size up to 16.} Speedup over the full model on Qwen3.5, median of three processes, on one RTX PRO 6000 Blackwell GPU. Park et al.'s kernels read about 37\% of routed-expert bytes. Park et al.'s whole-model runs use 99\% of GPU memory, and our reader and the full model use 95\%. At batch 32 and larger, Park et al.'s extra resident memory leaves too little room to run.}
\label{tab:park-speed}
\small
\setlength{\tabcolsep}{4.5pt}
\begin{tabular}{lrrrrrrrr}
\toprule
Batch size & 1 & 2 & 4 & 8 & 16 & 32 & 64 & 128 \\
\midrule
\multicolumn{9}{l}{\emph{MoE block}} \\
Our reader, 12.8\% of reads & 1.62 & 1.78 & 2.78 & 3.01 & 3.26 & 2.59 & 1.73 & 1.51 \\
Our reader, 17.8\% of reads & 1.55 & 1.71 & 2.49 & 2.73 & 2.68 & 2.19 & 1.50 & 1.29 \\
\citet{park2026intraexpert} & 0.67 & 0.92 & 1.28 & 1.65 & 1.98 & 2.09 & 2.08 & 1.90 \\
\midrule
\multicolumn{9}{l}{\emph{Whole model}} \\
Our reader, 12.8\% of reads & 1.19 & 1.24 & 1.52 & 1.71 & 2.08 & 1.97 & 1.48 & 1.31 \\
Our reader, 17.8\% of reads & 1.15 & 1.22 & 1.46 & 1.65 & 1.90 & 1.77 & 1.35 & 1.19 \\
\citet{park2026intraexpert} & 0.85 & 0.97 & 1.14 & 1.34 & 1.59 & -- & -- & -- \\
\bottomrule
\end{tabular}
\end{table}

\paragraph{Larger batches.} Tokens in a batch together select more channels, so the gain shrinks at large
batches. Whole-model speedup peaks at batch 8 or 16 (1.75$\times$ on Qwen3-Thinking, 1.74$\times$ on Gemma~4,
and 1.88$\times$ on Qwen3.5) and falls to 1.00$\times$, 1.07$\times$, and 1.24$\times$ at batch 128
(Table~\ref{tab:speed}).

\begin{figure}[tb]
\centering
\includegraphics[width=\linewidth]{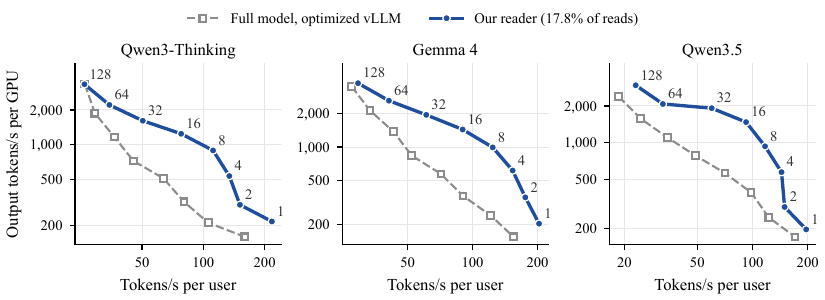}
\caption{\textbf{Our reader raises per-GPU throughput at the same per-user speed.}
Whole-model decoding of three post-trained models at 17.8\% of routed-expert reads, against optimized vLLM,
from Table~\ref{tab:speed} (128-token prompts, one RTX PRO 6000 Blackwell GPU). Each point is one batch size,
labeled for our reader. Per-user speed is the inverse of the time per output token, and per-GPU throughput is
the batch's total output rate.}
\label{fig:pareto}
\end{figure}

\paragraph{Serving trade-off.} Figure~\ref{fig:pareto} shows the trade-off between each user's speed and the
GPU's total throughput. At batch one, our reader generates 196 to 217 tokens per second per user. The full
model reaches this speed at no batch size; at batch one it generates 155 to 171 tokens per second. At the same
per-user speed, between 80 and 120 tokens per second, one GPU generates 3.3 to 4.4 times as many tokens per second
in total with our reader (3.3 to 4.3 at 100), interpolating log-linearly between measured batch sizes. The
full model must run smaller batches to stay that fast. At longer contexts, attention takes a larger share of
each step, and this advantage shrinks.

\paragraph{Comparison with Park et al.} \citet{park2026intraexpert} list their vLLM Triton kernels in their Appendix~B, so we time them against our reader on Qwen3.5, a model they also evaluate.
Park et al.'s kernels compute the gate projection of every routed expert in full with vLLM's fused-MoE kernel.
They keep the channels whose gate value is at least one global cutoff, and compute the up and down rows of those channels, reading the down rows from a transposed copy of the down weights.
We fixed one call in the listing that passed the cutoff where a block size belongs.
We fit the cutoff on decode states from our development prompts so that it keeps 6\% of routed channels, matching the 94\% routed sparsity Park et al.\ report at their 85\% total-sparsity target for this model; Park et al.'s kernels then read about 37\% of routed-expert bytes.
The transposed copy (20.0\,GiB) and the kernels' buffers (4.2\,GiB) add 24.2\,GiB of GPU memory, against 5.1\,GiB for our reader's sketch, recovery maps, and buffers.
Both versions run in the same harness as Table~\ref{tab:speed}, and our reader's speedups here agree with that table within 0.06$\times$.

Our reader at 17.8\% of reads is faster than Park et al.'s kernels in the MoE block up to batch 32, and Park et al.'s kernels are faster at batch 64 and 128 (Table~\ref{tab:park-speed}).
Park et al.'s sparse up and down kernels take as long as the full model's expert kernels at batch one (36.1 against 35.5\,$\mu$s per layer).
The full gate projection, the cutoff and compaction, and alignment add 23.7\,$\mu$s, so Park et al.'s MoE block runs slower than the full model at batch one.
Our reader is faster than Park et al.'s kernels in the whole model at every batch size up to 16.
Park et al.'s kernels do not fit in GPU memory at batch 32 and larger.
Park et al.'s threshold scores 67.6 at 38.0\% of reads on their zero-shot protocol for Qwen3.5 (ARC-Challenge, ARC-Easy, HellaSwag, WinoGrande, and TruthfulQA, with the prompt sparse too), close to the 67.5 they report.
Our reader scores 69.6 at 12.8\% of reads and 70.2 at 17.9\%, and the full model scores 71.2.
\end{document}